\documentclass{article} 
\PassOptionsToPackage{table}{xcolor}%
\usepackage{floatrow}
\usepackage{iclr2027_conference,times}

\usepackage{amsmath,amsfonts,bm}

\def\eqref#1{equation~\ref{#1}}

\def\1{\bm{1}}

\DeclareMathAlphabet{\mathsfit}{\encodingdefault}{\sfdefault}{m}{sl}
\SetMathAlphabet{\mathsfit}{bold}{\encodingdefault}{\sfdefault}{bx}{n}

\usepackage[
    colorlinks=true,
    citecolor=blue,
    linkcolor=purple,
    urlcolor=black
]{hyperref}
\usepackage{url}
\usepackage{amssymb}
\usepackage{amsmath} 
\usepackage{cleveref}
\usepackage{graphicx}
\usepackage{booktabs}
\usepackage{caption}
\usepackage{enumitem}
\usepackage{hyperref}
\usepackage{url}
\usepackage{amssymb}
\usepackage{amsmath}
\usepackage{amsfonts}%
\usepackage{algorithm}%
\usepackage{algpseudocode}%
\usepackage{cleveref}
\usepackage{textcomp}
\usepackage{amsthm}
\usepackage{multirow}
\usepackage{capt-of}
\usepackage{placeins}
\usepackage{wrapfig}
\usepackage{subcaption}

\usepackage{thmtools,thm-restate}
\usepackage{makecell}
\usepackage[table]{xcolor}
\usepackage{wrapfig}

\usepackage{xcolor}
\definecolor{repblue}{HTML}{1F6FB4}
\definecolor{colred}{HTML}{C0392B}
\definecolor{collapse}{HTML}{B03030}
\definecolor{hdr}{HTML}{333333}

\title{Noise-Space Reward Alignment for Few-Step Generative Models}%

\author{%
\textbf{
Woocheol Shin$^{1}$, \enspace
Taeyoung Yun$^{1}$, \enspace
Jiwon Sung$^{2}$, \enspace
Hyeongyu Kang$^{1}$, \enspace
Jeongmin Park$^{1}$, \enspace
}
\\
\textbf{
Jaewoo Lee$^{1,4}$, \enspace
Kyuil Sim$^{1}$, \enspace
Dohyun Kim$^{1}$, \enspace
Yutong He$^{3\dagger}$, \enspace
Jinkyoo Park$^{1,5\dagger}$
}
\\
$^1$KAIST, $^2$Stanford University, $^3$Carnegie Mellon University, $^4$MongooseAI, $^5$Omelet \\
{\small\texttt{\{woofe, 99yty, khg2000v, pjmin831\}@kaist.ac.kr}} \\
{\small\texttt{\{jaewoo, kyuil.sim, dyun0317, jinkyoo.park\}@kaist.ac.kr}} \\
{\small\texttt{jwsung@stanford.edu, yutonghe@andrew.cmu.edu}}
}

\iclrfinalcopy%

\begin{document}

\maketitle
\lhead{}
\ificlrfinal{\renewcommand{\thefootnote}{}\footnotetext{$^\dagger$Co-corresponding authors.\\\hspace*{1.8em}Project page: \url{https://shin-woocheol.github.io/npo/}}}\fi

\begin{abstract}
Few-step generative models produce high-quality samples in one or a few evaluations. 
However, steering them toward desired properties remains difficult, especially when only black-box reward evaluations are available. 
While one could apply approaches developed for iterative diffusion and flow-based models, such as direct RL fine-tuning or test-time scaling, we find that neither transfers well to the few-step regime: direct fine-tuning suffers from severe reward over-optimization, and test-time scaling sacrifices the efficiency advantage of few-step generation.
We propose Noise Policy Optimization (NPO), which explicitly targets the alignment problem for few-step generative models. 
The key idea is to learn an amortized flow policy over the initial noise space of the generator. 
The trained policy transports Gaussian noise to a reward-tilted noise distribution, whose pushforward distribution through the frozen generator is exactly the reward-tilted output distribution. 
By keeping the generator frozen, we can effectively utilize the knowledge encoded in the pretrained model, yielding a better reward-quality trade-off than direct fine-tuning. 
The amortized policy moves the search cost into training, so inference adds only a lightweight noise sampling step on top of the pretrained model.
To improve sample efficiency during training, we use ELBO-based likelihood estimation that requires only sampled endpoints rather than simulated trajectories.
Across image, language, and protein structure generation, NPO consistently achieves a better reward-quality trade-off than direct fine-tuning, and matches test-time search methods with \(16\text{ to }549\times\) lower inference cost.

\end{abstract}
\section{Introduction}
\label{sec:intro}

Recent advances in few-step generative modeling have enabled high-quality generation with only one or a few model evaluations~\citep{song2023consistency,boffi2024flow,chen2025sana,geng2026mean,hu2026elf,xie2025distilled}. However, one may need to align the generator toward desired properties in downstream applications, such as human preferences in image and text generation or structural properties in biological design.
Despite the growing importance of few-step generators, their reward alignment remains relatively underexplored.
Moreover, most of these downstream objectives are available only as scalar values whose gradients cannot be propagated through the generator due to non-differentiable rewards or discretization.
Together, these considerations motivate an important problem: how to align few-step generative models without access to reward gradients.

\begin{figure}[t]
\vspace{-5pt}
\centering
\includegraphics[width=\linewidth]{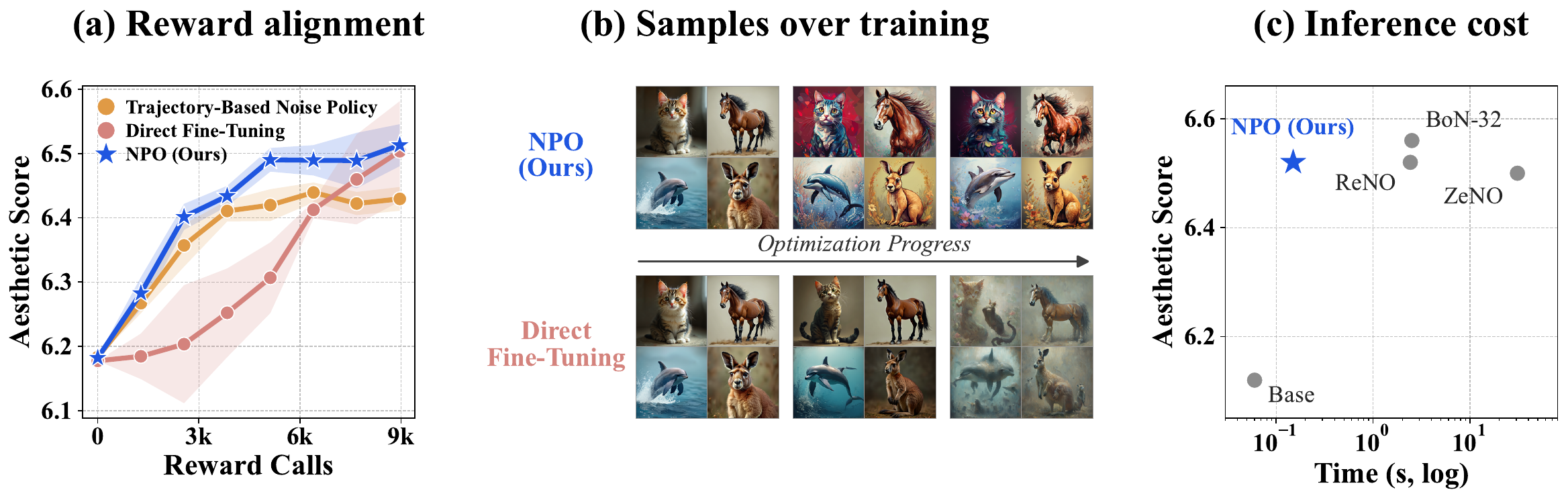}
\caption{\textbf{Motivation Figure.} (a) Reward learning curve of NPO, direct fine-tuning, and trajectory-based noise policy. (b) Generated samples over training for NPO and direct fine-tuning. (c) Comparison with test-time scaling methods. As shown in the figure, NPO achieves a better reward-quality trade-off than baselines while maintaining an inference cost comparable to the pretrained model.}
\label{fig:motivation}
\vspace{-10pt}
\end{figure}

A straightforward way is to directly fine-tune the generator using reinforcement learning (RL) methods developed for iterative diffusion and flow-based models~\citep{fan2023dpok,black2024training,liu2025flowgrpo}. However, we find that direct fine-tuning transfers poorly to the few-step regime. As demonstrated in \Cref{fig:motivation}, direct methods either make limited progress or suffer from severe reward over-optimization that degrades sample quality. 
We hypothesize that directly changing the parameters of a few-step generative model is prone to losing the knowledge encoded during pretraining. 

A promising alternative is to keep the generator frozen and optimize its input noise at test time~\citep{eyring2024reno,kim2026gradient,yoon2026psi,min2026origen}. While searching the noise through test-time scaling can effectively improve the target reward without modifying the pretrained generator, the process must be repeated from scratch for every generated sample. 
This incurs substantial inference cost and sacrifices the efficiency advantage of few-step generation.

We propose Noise Policy Optimization (NPO), which learns an amortized flow policy over the input noise space of a frozen few-step generator. 
We use the identity that the problem of sampling from the reward-tilted output distribution can be reduced to sampling from the reward-tilted input noise distribution when the sampling path is deterministic.
NPO learns this transformation with a flow policy that transports the standard Gaussian distribution toward the reward-tilted input noise distribution. 
By keeping the generator frozen, NPO preserves the knowledge in the pretrained model. By amortizing the noise transformation, it moves the cost of test-time search to training.
To enable more efficient training, we use ELBO-based likelihood estimation from sampled noise endpoints instead of using conventional trajectory-based likelihood estimation~\citep{black2024training,liu2025flowgrpo}. We observe that it drastically improves the sample efficiency of our method, as depicted in \Cref{fig:motivation}.

We evaluate NPO across image, language, and protein structure generation tasks, covering a diverse range of one- and few-step generators and reward functions. Compared to direct generator fine-tuning, NPO consistently achieves a better trade-off between improving reward and preserving sample quality and diversity. Relative to test-time noise optimization, NPO achieves comparable or higher rewards with \(16 \text{ to }549\times\) lower inference cost by learning an amortized policy. In comparison to trajectory-based noise-policy training, NPO reaches the same reward levels with fewer reward evaluations. 
Together, these results demonstrate that NPO efficiently aligns few-step generative models using only black-box rewards, without sacrificing sample quality or inference efficiency.

\section{Related Work}
\label{sec:related}

\textbf{Few-step generative models.}
Recent advances have enabled high-quality generation with only one or a few model evaluations.
Consistency models learn trajectory-consistent mappings~\citep{song2023consistency, song2024improved, geng2025consistency, luo2023latent}, while distillation-based methods compress multi-step generators using objectives such as adversarial training or distribution matching~\citep{sauer2024adversarial, yin2024one, luo2023diff}.
Flow-map-based and related approaches learn finite-time transport between noise levels or directly from noise to data~\citep{kim2024consistency, frans2025one, geng2026mean, boffi2024flow, boffi2026build}.
While these methods substantially reduce sampling cost, aligning them with downstream rewards remains underexplored.

\textbf{Reward alignment of generative models.}
Generative models can be aligned to downstream objectives by reward backpropagation~\citep{prabhudesai2023aligning, xu2023imagereward, kang2026diffusion}, RL-based policy optimization~\citep{black2024training, fan2023dpok, zhang2024confronting, liu2025flowgrpo}, incorporating rewards during distillation~\citep{duan2026accelerating}, or using variational inference~\citep{lee2026diffusion}. Recent work has also studied alignment of few-step generators through policy optimization for consistency models~\citep{oertell2024rl}, sample-based variational inference~\citep{lee2026aligning}, and flow-map-specific approaches~\citep{mammadov2026variational, potaptchik2026meta}. 
Although promising, these approaches either directly modify the pretrained generator, which can lead to reward hacking, or rely on model-specific structures. In contrast, NPO keeps the pretrained generator frozen to preserve sample quality and applies across diverse few-step generators.

\textbf{Initial-noise optimization.}
For deterministic generators, alignment can alternatively be performed by optimizing the input noise while keeping the generator fixed.
Test-time methods optimize the initial noise for each sample using reward gradients~\citep{eyring2024reno, min2026origen, yoon2026psi} or gradient-free optimization~\citep{ma2025inference, kim2026gradient}, but incur repeated inference-time search.
Amortized approaches instead learn reusable noise distributions or policies, including Gaussian noise policies~\citep{chen2024find, miao2025noiseppo}, reward-gradient-based transformations~\citep{eyring2025noise, pmlr-v305-wagenmaker25a}, and diffusion-based noise samplers~\citep{venkatraman2025outsourced}.
ELEGANT~\citep{uehara2024fine} controls continuous-time diffusion dynamics but does not directly applicable to the few-step generators considered in our paper.
NPO similarly amortizes noise-space optimization, while supporting black-box reward evaluations with an expressive flow policy trained using endpoint forward-noising updates to improve sample efficiency.

\section{Problem Setup \& Background}
\label{sec:background}

In this section, we study reward alignment of pretrained few-step generative models. Given a generator and a scalar reward function, our goal is to steer generation toward high-reward outputs without assuming access to reward gradients. We first formalize the desired reward-tilted output distribution and review trajectory-based policy optimization. 

\subsection{Problem Setup and Reward-Tilted Output Distribution}
\label{sec:bg-target}

Let $g_\theta$ denote a pretrained few-step generator that learns to transport from a source distribution $p_1$ to a target distribution $p_{\mathrm{data}}=p_0$. We denote the input $x_1\sim p_1=\mathcal N(0,I)$, the output $x=g_\theta(x_1)$, and the output distribution $p_{\mathrm{base}}=g_{\theta\#}p_1$. Additionally, we have access to a scalar reward $r:\mathcal X\rightarrow\mathbb R$, which can be evaluated on generated samples but does not need to be differentiable.

Reward alignment seeks to improve the expected reward while limiting deviation from the pretrained distribution. A standard KL-regularized objective and its optimal output distribution are
\begin{equation}
    \label{eq:tilted-output}
    \max_p
    \left\{
    \mathbb E_{x\sim p}[r(x)]
    -
    \alpha D_{\mathrm{KL}}(p\Vert p_{\mathrm{base}})
    \right\},
    \qquad
    p^\star(x)
    =
    \frac{1}{Z}
    p_{\mathrm{base}}(x)
    \exp\!\left(\frac{r(x)}{\alpha}\right),
\end{equation}
where $\alpha>0$ controls the regularization strength and $Z$ is the normalizing constant. We call $p^\star$ the \emph{reward-tilted output distribution}, which reweights the pretrained distribution toward high-reward outputs. Thus, reward alignment can equivalently be viewed as learning to sample from $p^\star$. A detailed derivation is provided in \Cref{app:kl-solution}.

\subsection{Trajectory-Based Policy Optimization}
\label{sec:bg-trajectory}

Let $p_\phi$ denote a trainable output distribution. When reward gradients are unavailable, the reward term in \Cref{eq:tilted-output} can be optimized using the score-function, or likelihood-ratio, identity
\begin{equation}
    \label{eq:score-function-pg}
    \nabla_\phi
    \mathbb E_{x\sim p_\phi}[r(x)]
    =
    \mathbb E_{x\sim p_\phi}
    \left[
    r(x)\nabla_\phi\log p_\phi(x)
    \right].
\end{equation}
Hence, policy-gradient optimization requires estimating the endpoint likelihood $p_\phi(x)$.

For iterative diffusion and flow models, this likelihood is commonly estimated from simulated reverse trajectories. Given a simulated trajectory $x_{\tau_K},\hat{x}_{\tau_{K-1}},\ldots,\hat{x}_{\tau_0}$ where $0=\tau_0<\tau_1<\cdots<\tau_K=1$, $x_{\tau_K} = x_1 \sim p_1$ is the initial noise, $\hat{x}_{\tau_k}$ is the $k$-th step intermediate sample and $x=\hat{x}_0$ is the final generated sample, the endpoint log likelihood can be estimated as
\begin{equation}
    \label{eq:trajectory-likelihood}
    \log p_\phi(x)
    \approx
    \sum_{k=1}^{K-1}
    \log p_\phi(\hat{x}_{\tau_{k-1}}\mid \hat{x}_{\tau_k})
    + \log p_\phi(\hat{x}_{\tau_{K-1}}\mid x_{\tau_{K}}) +
    C(x)
\end{equation}
where $C(x)$ is a $\phi$-independent constant.

Prior works such as DDPO~\citep{black2024training} and Flow-GRPO~\citep{liu2025flowgrpo} use this trajectory-based likelihood estimation to optimize the reward objective. While this construction is tractable for stochastic diffusion samplers because their discretized reverse transitions are Gaussian, it does not naturally apply to deterministic ODE and flow-map samplers. Existing approaches therefore resort to surrogate constructions, including introducing additional stochasticity into the sampling process~\citep{liu2025flowgrpo}, computing likelihoods through divergence estimators~\citep{lipman2023flow}, and learning auxiliary models for likelihood estimation~\citep{ai2026joint}. However, these techniques are prohibitively expensive and unstable~\citep{ouyang2026few}, or fundamentally change the sampling process and cause discrepancies between reward alignment and base model pretraining.

\section{Noise Policy Optimization}
\label{sec:method}

To address these challenges, we propose Noise Policy Optimization (NPO). NPO learns an amortized flow policy over the input noise of a frozen generator. We first show that learning the reward-tilted noise distribution is sufficient to realize the reward-tilted output distribution through the frozen generator. Then, we propose an ELBO-based objective that approximates the target noise distribution with the flow policy in the noise space.

\begin{figure}[t]
\centering
\includegraphics[width=\linewidth]{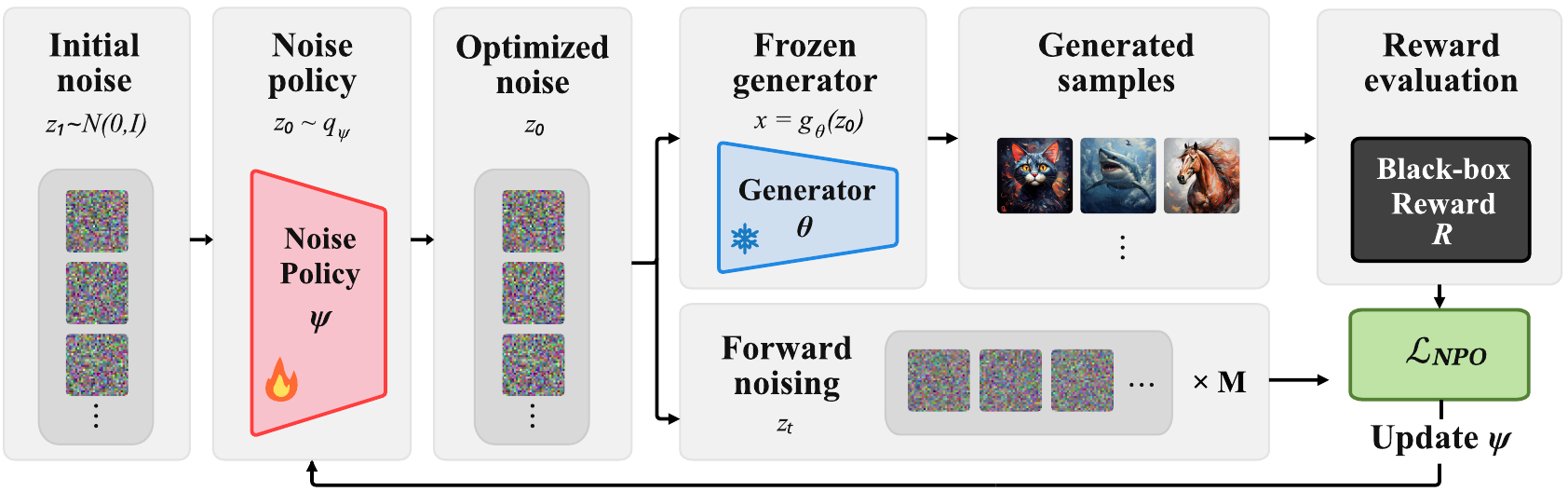}
\caption{\textbf{Overview of NPO.}
NPO trains a noise policy $q_\psi$ that maps Gaussian noise $z_1$ to optimized noise $z_0$ for a frozen generator $g_\theta$.
The policy is updated with $\mathcal{L}_{\mathrm{NPO}}$, computed from the black-box reward of $g_\theta(z_0)$ and $M$ forward-noised points $z_t$ of each $z_0$, given by forward-noising pairs $(t,\epsilon)$.
}
\label{fig:npo-overview}
\end{figure}

\subsection{Noise-Space Representation of the Target}
\label{sec:bg-noise-space}

Rather than modifying $g_\theta$ to realize the reward-tilted output distribution $p^\star$, we can instead change its input distribution while keeping the generator fixed. For notational simplicity, we denote the noise space as $\mathcal Z$ and the initial noise as $z\sim q_{\mathrm{base}}=p_1=\mathcal N(0,I)$.
Following prior noise-space formulations~\citep{venkatraman2025outsourced, eyring2025noise}, we define the \emph{reward-tilted noise distribution}
\begin{equation}
    \label{eq:tilted-noise}
    q^\star(z)
    =
    \frac{1}{Z}
    q_{\mathrm{base}}(z)
    \exp\!\left(
    \frac{r(g_\theta(z))}{\alpha}
    \right).
\end{equation}
Then, $g_{\theta\#}q^\star=p^\star$.
In other words, the pushforward distribution of  $q^\star$ is exactly the reward-tilted output distribution $p^\star$. Therefore, sampling from $p^\star$ can be reduced to sampling $z\sim q^\star$ and passing the sampled noise through the deterministic mapping $g_\theta$. We provide the proof in \Cref{app:proof-pushforward}.

\subsection{Flow Policies and Forward Noising}
\label{sec:bg-flow}
Motivated by the insight above, we propose to learn an additional model to transform the original Gaussian noise distribution into the reward-tilted noise distribution.
We parameterize a learned noise distribution $q_\psi$ using another continuous flow. In particular, we aim to learn a velocity field $v_\psi(z_t,t)$ that defines an ODE $dz_t/dt=v_\psi(z_t,t)$, whose integration from $t=1$ to $t=0$ transports a Gaussian source $z_1\sim q_1 = p_1 = \mathcal{N}(0,I)$ to an endpoint $z_0 \sim q_\psi \approx q_0 = q^*$. 

For an endpoint $z_0$, we construct the Gaussian forward-noising interpolation $z_t=a_tz_0+\sigma_t\epsilon$, where $\epsilon\sim\mathcal N(0,I)$, $(a_0,\sigma_0)=(1,0)$, and $(a_1,\sigma_1)=(0,1)$. The corresponding conditional target velocity is $u_t=\dot a_tz_0+\dot\sigma_t\epsilon$, where dots denote derivatives with respect to $t$.

\subsection{Noise-Space Alignment Objective}
\label{sec:method-objective}
To match $q_\psi$ to the target reward-tilted noise distribution $q^*$, we minimize the reverse KL divergence to $q^*$, which is equivalent to KL-regularized reward maximization in noise space: 
\begin{equation}
    \label{eq:noise-objective}
    \arg\min_\psi
    D_{\mathrm{KL}}(q_\psi\Vert q^\star)
    =
    \arg\min_\psi
    \left\{
    \alpha D_{\mathrm{KL}}(q_\psi\Vert q_{\mathrm{base}})
    -
    \mathbb E_{z\sim q_\psi}
    [r(g_\theta(z))]
    \right\}.
\end{equation}
Thus, our objective maximizes the reward while regularizing the learned noise distribution $q_\psi$ toward the initial noise distribution $q_{\mathrm{base}}$. To implement this reference regularization, we introduce a fixed reference velocity field $v_{\mathrm{ref}}(z_t,t)
    =
    \frac{
    a_t\dot a_t+\sigma_t\dot\sigma_t
    }{
    a_t^2+\sigma_t^2
    }
    z_t$, whose source and endpoint distributions are both $q_{\mathrm{base}}$.  
The forms of $v_{ref}$ under specific interpolation schedules are provided in \Cref{app:reference-field}.

\subsection{ELBO-Based Policy Optimization}
\label{sec:method-elbo}
The objective in \Cref{eq:noise-objective} can be well optimized via policy gradients. As a result, it is crucial to obtain a reliable estimate of the endpoint likelihood $q_\psi(z_0)$. Prior works show that ELBO-based estimators provide substantially more efficient optimization than trajectory-based alternatives~\citep{xue2025advantage,choi2026rethinking}. We therefore use an ELBO-based estimator that operates on sampled endpoints and avoids dependence on simulated rollout trajectories.

For endpoint $z_0$, defining the regression score on a forward-noising pair $(t,\epsilon)$, with $t\sim\mathcal U(0,1)$ and $\epsilon\sim\mathcal N(0,I)$, as $\ell_\psi(z_0;t,\epsilon)=w(t)\Vert v_\psi(z_t,t)-u_t\Vert_2^2$, we estimate the endpoint log likelihood as
\begin{equation}
    \label{eq:elbo-likelihood}
    \log q_\psi(z_0)
    \approx
    -
    \mathbb E_{t,\epsilon}
    \left[
    \ell_\psi(z_0;t,\epsilon)
    \right]
    +
    c(z_0)
\end{equation}
where $c(z_0)$ is a $\psi$-independent constant. Weighting and estimation details are in \Cref{app:elbo-weighting}.

In practice, we maintain an old policy $q_{\psi_{\mathrm{old}}}$ for sampling endpoints. 
Following ~\cite{xue2025advantage}, we estimate the importance ratio between the current and old policies on the same forward-noising pair as $\hat\rho_\psi = \exp(\ell_{\psi_{\mathrm{old}}}-\ell_\psi)$, which serves as a heuristic surrogate.
We use the advantage $A=r(g_\theta(z_0))-\bar r$, where $\bar r$ is the group-mean reward. Following prior work~\cite{choi2026rethinking}, we approximate the KL regularization term by the squared deviation between $v_\psi$ and $v_{\mathrm{ref}}$. Additional details are in \Cref{app:reference-regularization}. Combining these terms, our final loss is
\begin{equation}
    \label{eq:npo-loss}
    \mathcal L_{\mathrm{NPO}}(\psi)
    =
    \mathbb E
    \left[
    \operatorname{sg}
    \left(
    A\widehat\rho_\psi
    \right)w(t)
    \left\Vert
    v_\psi(z_t,t)-u_t
    \right\Vert_2^2
    +
    \lambda_{\mathrm{ref}}
    \left\Vert
    v_\psi(z_t,t)-v_{\mathrm{ref}}(z_t,t)
    \right\Vert_2^2
    \right].
\end{equation}

Here, $\operatorname{sg}$ denotes stop-gradient and $\lambda_{\mathrm{ref}}>0$ controls the strength of the reference regularization. We summarize the overall NPO training procedure in \Cref{alg:npo} and provide additional methodological details in \Cref{app:policy-optimization}.

\begin{algorithm}[t]
\caption{Noise Policy Optimization}
\label{alg:npo}
\begin{algorithmic}[1]
\Require Frozen generator $g_\theta$; reward $r$; noise policy $q_\psi$ with velocity field $v_\psi$;
         reference field $v_{\mathrm{ref}}$; number of forward-noising pairs $M$; training rounds $N$; EMA decay $\gamma$
\State Initialize $v_\psi \leftarrow v_{\mathrm{ref}}$, $\psi_{\mathrm{old}} \leftarrow \psi$
\For{$k = 1, \dots, N$}
    \State Sample a batch of $z_1 \sim q_{\mathrm{base}}$ and integrate $v_{\psi_{\mathrm{old}}}$ to obtain endpoints $z_0$
    \State Compute rewards $r(g_\theta(z_0))$ and advantages $A = r(g_\theta(z_0)) - \bar r$
    \State Sample $M$ forward-noising pairs $(t, \epsilon)$ per endpoint
    \State Update $\psi$ by minimizing $\mathcal L_{\mathrm{NPO}}(\psi)$ from \Cref{eq:npo-loss}
    \State $\psi_{\mathrm{old}} \leftarrow \gamma\,\psi_{\mathrm{old}} + (1 - \gamma)\,\psi$
\EndFor
\end{algorithmic}
\end{algorithm}

\section{Experiments}
\label{sec:exp}

We evaluate NPO across image, language, and protein structure generation, covering diverse one- and few-step generators and reward functions. Our experiments aim to address the following three questions: (1) Does NPO achieve a better reward-quality trade-off than direct generator fine-tuning? (2) Does NPO achieve competitive performance compared to test-time methods at lower inference cost? (3) Does NPO learn an amortized policy faster than trajectory-based approaches? 

\begin{figure}[t]
\vspace{-12pt}
\centering
\includegraphics[width=0.98\linewidth]{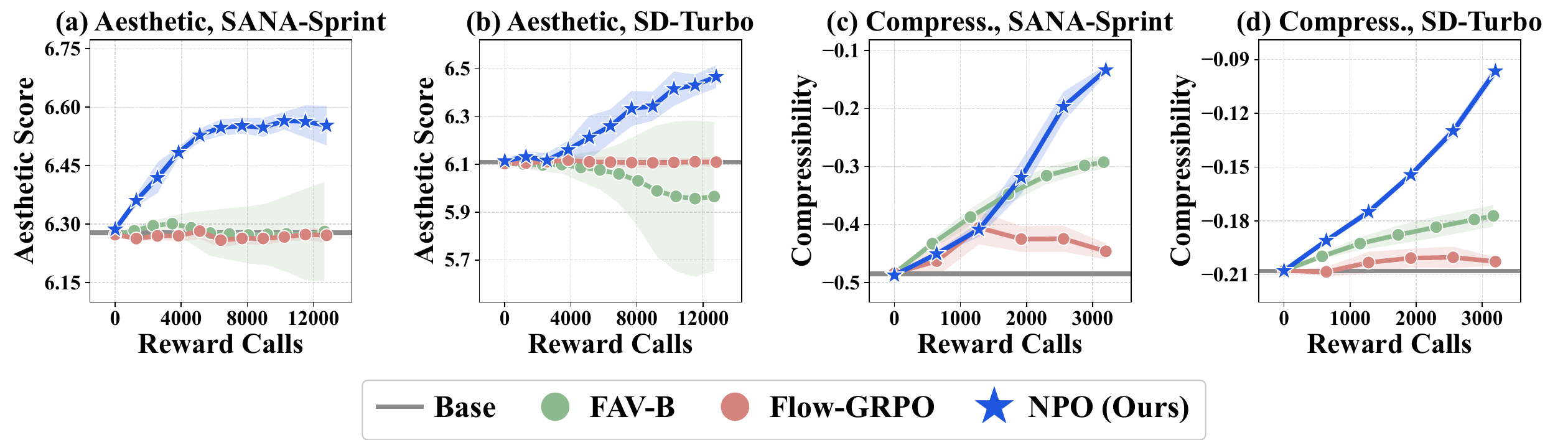}
\caption{Reward learning curve of NPO and direct fine-tuning baselines on 1-step generation. NPO efficiently learns to generate high-rewarding samples compared to the baselines.}
\label{fig:image_target_1step}
\vspace{-5pt}
\end{figure}

\begin{figure}[t]
\vspace{-9pt}
\centering
\includegraphics[width=0.98\linewidth]{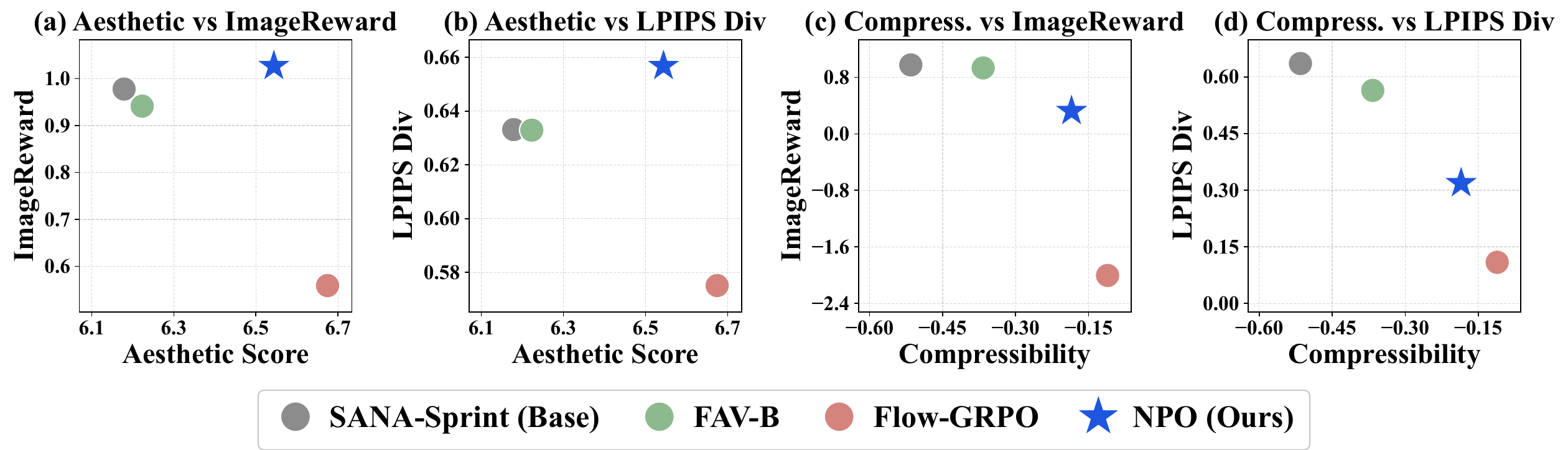}
\caption{Reward-quality trade-off of NPO and direct fine-tuning baselines on SANA-Sprint with 4-step sampling. NPO achieves a better trade-off than the baselines.}
\vspace{-5pt}
\label{fig:image_pareto_4step}
\end{figure}

\subsection{Text-to-Image Generation}
\label{sec:exp-image}
We first evaluate NPO on text-to-image generation tasks. For pretrained few-step generators, we use SD-Turbo~\citep{sauer2024adversarial} and SANA-Sprint~\citep{chen2025sana}, which support one- and four-step image generation. We provide additional experimental details in \Cref{app:baselines}.

\subsubsection{Experimental Setup}
\label{sec:exp-image-setup}

\textbf{Rewards and Metrics.} 
We optimize two target reward functions: the LAION aesthetic predictor~\citep{schuhmann2022laion} and JPEG compressibility~\citep{black2024training}, without using reward gradients.
To measure reward over-optimization, we report ImageReward~\citep{xu2023imagereward} for semantic fidelity and LPIPS diversity~\citep{zhang2018unreasonable} for sample diversity.

\textbf{Baselines.} 
We compare NPO against three categories of alignment methods. Implementation details for all baselines are provided in \Cref{app:baseline-impl}.

\textbf{(1) Direct generator fine-tuning.}
Flow-GRPO~\citep{liu2025flowgrpo} applies policy-gradient updates directly to the generator using stochastic transition likelihoods.
FAV-B~\citep{lee2026aligning} fine-tunes the generator using amortized SVGD with zeroth-order gradient estimation from reward evaluations.

\textbf{(2) Test-time noise optimization.}
Best-of-$N$~\citep{karthik2023if} samples $N$ candidate noises and selects the highest-reward output.
ReNO~\citep{eyring2024reno} and ZeNO~\citep{kim2026gradient} perform gradient-based and gradient-free noise optimization at test-time, respectively.

\textbf{(3) Noise-policy learning.}
ODS~\citep{venkatraman2025outsourced} learns a score model in noise space using a trajectory-balance objective. We additionally implement Noise Flow-GRPO, a variant of our method that instead uses trajectory-based likelihood estimation.

\subsubsection{Results}
\label{sec:exp-image-results}
\textbf{NPO improves reward while preserving sample quality.}
\Cref{fig:image_target_1step} compares NPO with direct generator fine-tuning under one-step sampling.
Across both pretrained generators, NPO consistently improves the target rewards, whereas Flow-GRPO and FAV-B show limited or no improvement.
Under four-step sampling, \Cref{fig:image_pareto_4step} shows that Flow-GRPO improves the target rewards, but ImageReward and LPIPS diversity decrease substantially, indicating severe reward over-optimization.
In contrast, NPO improves the target rewards while better preserving semantic fidelity and diversity.
By keeping the pretrained generator frozen and aligning in noise space, NPO achieves a more favorable reward-quality trade-off.
Additional qualitative results are provided in \Cref{app:qualitative-Image}.

\par\medskip
\noindent
\begin{minipage}{\linewidth}
    \begin{minipage}[t]{0.48\linewidth}
        \vspace{0pt}
        \textbf{NPO preserves fast inference by amortizing test-time search.}
        \Cref{tab:sdturbo_test_time} compares NPO with test-time noise optimization in terms of the inference speed. NPO achieves comparable reward while reducing inference cost approximately \(16\)–\(205\times\) relative to the test-time baselines. By amortizing test-time search into a reusable flow policy, NPO achieves comparable alignment while preserving the fast inference of few-step generation.
    \end{minipage}\hfill
    \begin{minipage}[t]{0.48\linewidth}
        \vspace{0pt}
        \centering
        \resizebox{\linewidth}{!}{%
            
\begin{tabular}{lrrrr}
\toprule
Method & Aesthetic & NFE & Time (s) & Rel. cost \\
\midrule
SD-Turbo (Base) & $6.09 \pm 0.01$ & 1   & 0.06  & $1.0\times$ \\
BoN-32         & $6.57 \pm 0.01$ & 32  & 2.51  & $44.2\times$ \\
ReNO           & $6.52 \pm 0.01$ & 13  & 2.42  & $42.6\times$ \\
ZeNO           & $6.50 \pm 0.00$ & 482 & 30.78 & $542.8\times$ \\
\midrule
\textbf{NPO (Ours)} & $6.46 \pm 0.04$ & 1 & 0.15 & $2.6\times$ \\
\bottomrule
\end{tabular}%
        }
        \captionof{table}{Comparison of NPO with test-time search methods on SD-Turbo. NPO achieves comparable aesthetic scores with a single generator evaluation.}
        \label{tab:sdturbo_test_time}
    \end{minipage}
\end{minipage}
\par\medskip

\begin{figure}[t]
\vspace{-10pt}
\centering
\includegraphics[width=\linewidth]{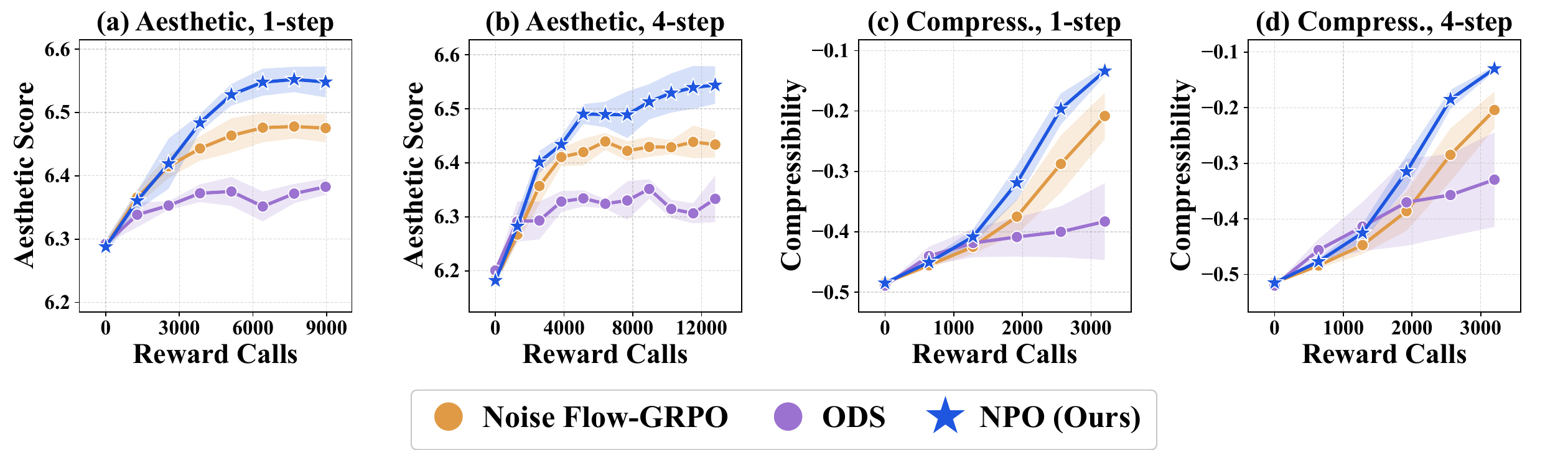}
\caption{Reward learning curve of NPO and noise-policy learning baselines on few-step generation. NPO efficiently learns to generate high-rewarding samples compared to the baselines.}
\label{fig:image_noise_efficiency}
\end{figure}

\textbf{NPO improves training efficiency with ELBO-based likelihood estimation.} 
\Cref{fig:image_noise_efficiency} compares NPO with Noise Flow-GRPO and ODS across one- and four-step sampling. NPO reaches comparable reward levels with fewer reward evaluations across all settings. Since Noise Flow-GRPO uses the same amortized flow policy, this controlled experiment reveals that avoiding dependence on stochastic rollout trajectories yields a more efficient optimization signal. NPO also outperforms ODS, indicating that the advantage extends to a broader trajectory-based noise-policy baseline.

\subsection{Language Generation}
\label{sec:exp-language}
Recent work has begun to explore few-step language generation with continuous-flow models~\citep{hu2026elf,lee2026flow}. To this end, we evaluate NPO on alignment of few-step language generation models. Additional experimental details on the language task are provided in \Cref{app:lang-details}.

\subsubsection{Experimental Setup}
\label{sec:exp-language-setup}
We use the progressively distilled four-step variant of ELF~\citep{hu2026elf} as the pretrained generator.
We introduce a DeBERTa-v3-large NLI model~\citep{he2021debertav3}, which is fine-tuned on MNLI~\citep{williams2018broad}, to define the target reward as the entailment probability for the hypothesis “This text is about [topic].”.
We steer generation toward two topics, politics and science. To evaluate generation quality beyond the target reward, we report GPT-2 Large perplexity~\citep{radford2019language} for fluency and unigram entropy~\citep{lee2026flow} for diversity. We compare against the same  baselines as in the image experiments. Note that as the reward is scored on the text only after sampling, the gradient is not straightforwardly available.

\begin{figure}[t]
\vspace{-5pt}
\centering
\includegraphics[width=\linewidth]{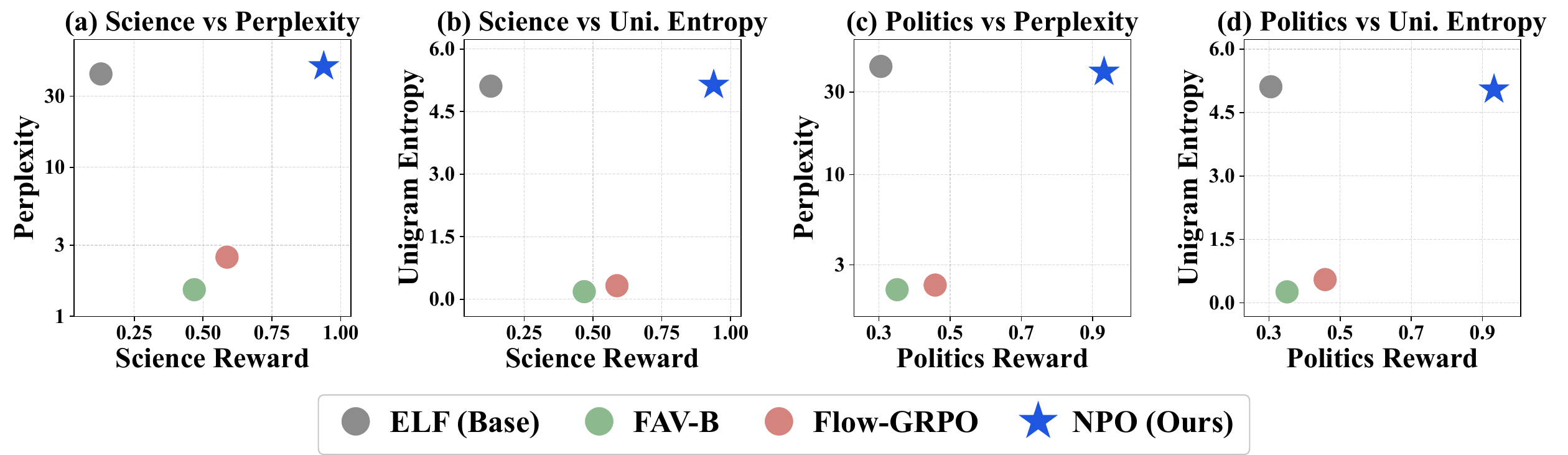}
\caption{Reward-quality trade-off of NPO against direct fine-tuning on the topic-steering task. NPO maintains perplexity and entropy comparable to those of the base model while achieving higher reward than the baselines.}
\label{fig:language_pareto}
\vspace{-5pt}
\end{figure}

\vspace{-14pt}
\subsubsection{Results}
\label{sec:exp-language-results}

\textbf{NPO improves reward while preserving sample quality.} 
\Cref{fig:language_pareto} compares NPO with direct fine-tuning methods. The direct fine-tuning baselines achieve only modest improvements in topic reward while degrading generation quality. The simultaneous decrease in both perplexity and unigram entropy indicates degenerate repetition rather than improved fluency, as further illustrated by the qualitative samples in \Cref{app:qualitative-lang}. In contrast, NPO achieves higher topic rewards while keeping both perplexity and unigram entropy close to those of the frozen generator. These results show that NPO achieves a better reward–sample quality trade-off than direct fine-tuning.

\begin{figure}[t]
    \vspace{9pt}
    \centering
    \begin{subfigure}[b]{0.45\textwidth}
        \centering
        \resizebox{\linewidth}{!}{%

\begin{tabular}{lrrrr}
    \toprule
    Method & Reward & NFE & Time (s) & Rel. cost \\
    \midrule
    ELF (Base) & $0.14 \pm 0.02$ & 4     & 0.03  & $1.0\times$ \\
    BoN-64     & $0.96 \pm 0.01$ & 256   & 5.87  & $218.2\times$ \\
    ZeNO       & $0.54 \pm 0.03$ & 1,928 & 38.41 & $1427.8\times$ \\
    \midrule
    \textbf{NPO (Ours)} & $0.94 \pm 0.01$ & 4 & 0.07 & $2.5\times$ \\
    \bottomrule
\end{tabular}%
        }
        \vspace{20pt}
        \caption{Comparison with test-time search methods.}
        \label{fig:lang-testtime}
    \end{subfigure}\hfill
    \begin{subfigure}[b]{0.5\textwidth}
        \centering
        \includegraphics[width=\linewidth]{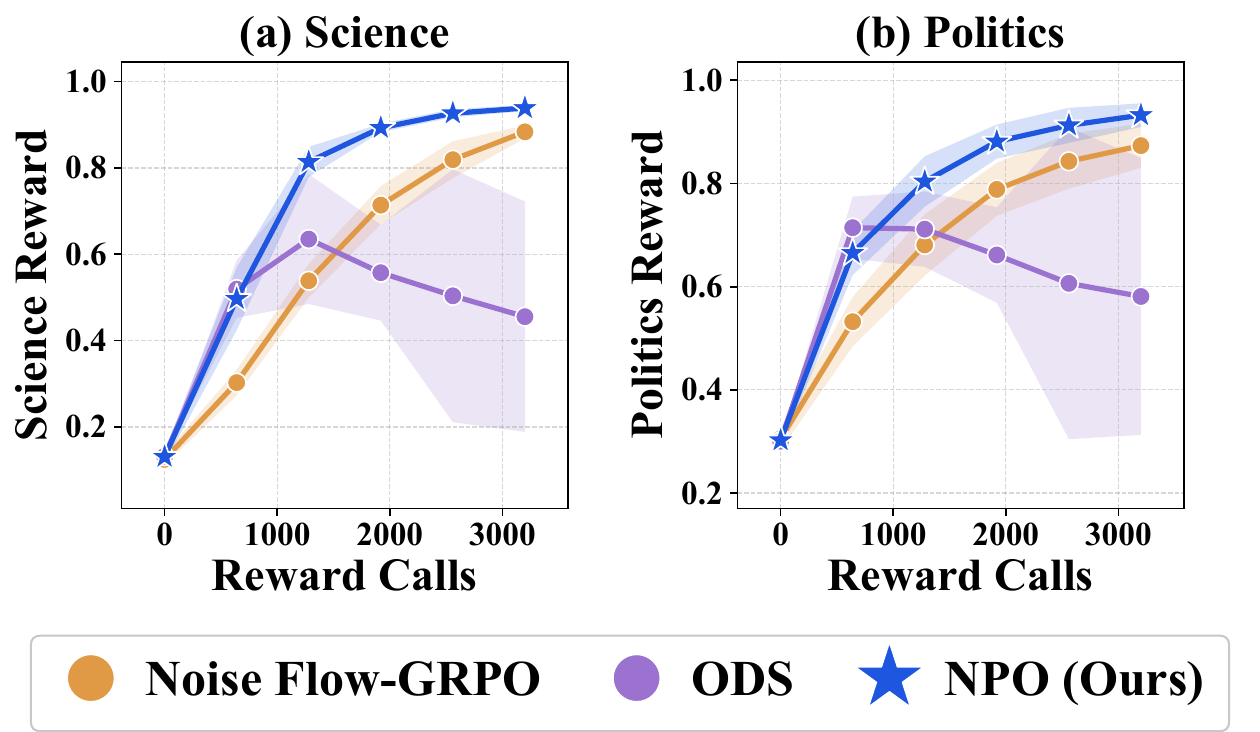}
        \caption{Learning curves of noise-policy learning methods.}
        \label{fig:lang_noise_efficiency}
    \end{subfigure}
    \vspace{-3pt}
    \caption{Results on the topic-steering task with ELF. (a) NPO achieves comparable reward to test-time search methods with only four generator evaluations. (b) NPO learns to generate high-reward samples more efficiently than noise-policy learning baselines.}
    \label{fig:lang-results}
\end{figure}

\textbf{NPO preserves fast inference by amortizing test-time search.}
\Cref{fig:lang-testtime} compares NPO with test-time search methods.
NPO achieves competitive scores while generating samples \(84\text{ to }549\times\) faster.
This shows that NPO effectively amortizes the cost of test-time search into training.

\textbf{NPO improves training efficiency with ELBO-based likelihood estimation.}
\Cref{fig:lang_noise_efficiency} compares the learning curves of NPO and noise-policy learning baselines.
Across both topics, NPO achieves higher reward than the baselines under the same reward-evaluation budget.
This shows that ELBO-based likelihood estimation enables more efficient noise-policy optimization.

\subsection{Protein Structure Generation}
\label{sec:exp-bio}
Several works have been developed to utilize few-step generative models for protein backbone generation~\citep{xie2025distilled, woo2026riemannian}. However, protein generative models tend to under-sample \(\beta\)-sheet-rich structures, limiting the diversity of generated secondary structures~\citep{venkatraman2025outsourced,su2025iterative}. Therefore, we evaluate NPO on protein structure tasks to promote diverse secondary structures with increased \(\beta\)-sheet content. Additional details are provided in \Cref{app:bio-details}.

\subsubsection{Experimental Setup}
\label{sec:exp-bio-setup}

We use the five-step protein backbone generator of~\cite{xie2025distilled} as a base pretrained model. Following \cite{venkatraman2025outsourced}, we use the secondary-structure diversity reward
\begin{equation}
    r_{\mathrm{div}}(x)
    =
    \frac{e^{-1.2}\, w^\top p}{1.2 - H[p]},
    \qquad
    w = [1, 2, 0.5],
\end{equation}
where $p = [p_\alpha, p_\beta, p_c]$ denotes the proportions of residues in $\alpha$-helices, $\beta$-sheets, and coils, respectively, and $H[p]$ is the entropy of $p$. 
We optimize and report $\log r_{\mathrm{div}}(x)$, termed SS composition, which favors diverse secondary-structure compositions with additional emphasis on $\beta$-sheet content.
To evaluate sample quality, we report designability and pairwise TM-score. Designability measures whether generated backbones can be self-consistently realized by designed amino-acid sequences, serving as a measure of plausibility. Pairwise TM-score measures structural similarity among backbones, with lower values indicating higher diversity.

\subsubsection{Results}
\label{sec:exp-bio-results}

\begin{figure}[t]
\vspace{-10pt}
\centering
\includegraphics[width=0.88\linewidth]{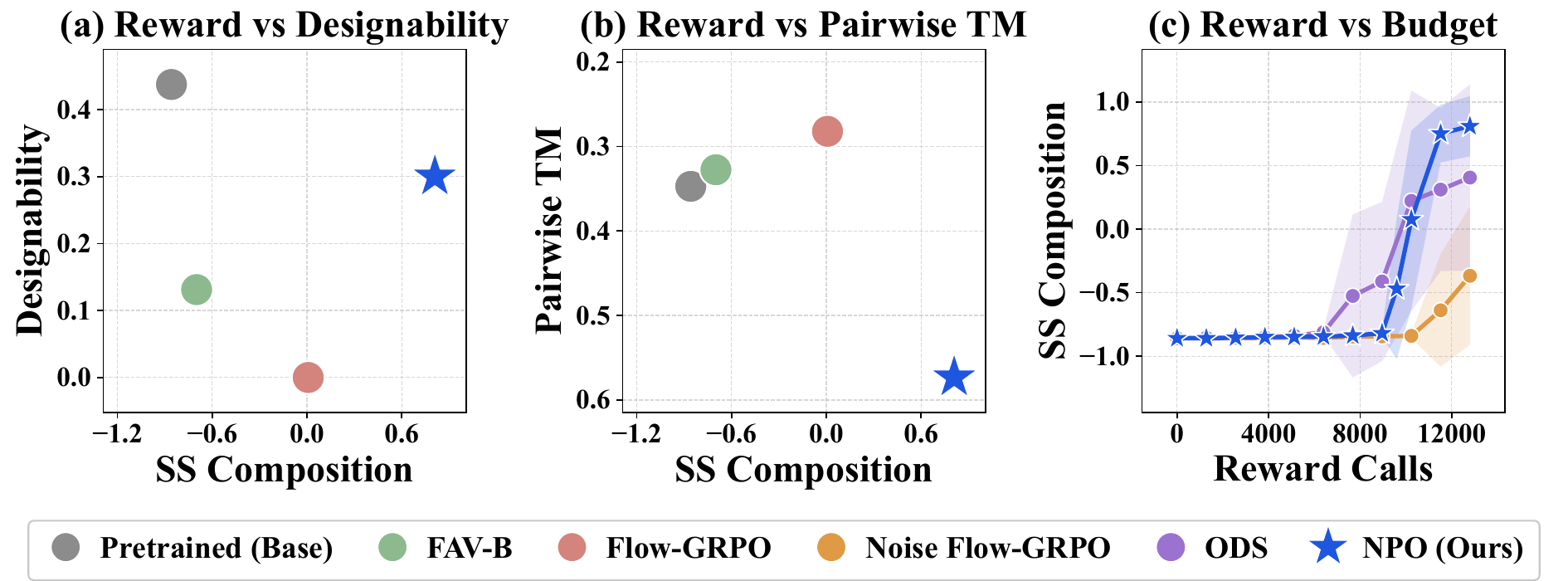}
\caption{Comparison of NPO against baselines on biological structure generation. NPO achieves a better reward-quality trade-off and training efficiency than the baselines.}
\label{fig:bio_figure}
\end{figure}

\textbf{NPO improves reward while preserving sample quality.} Panels (a) and (b) of \Cref{fig:bio_figure} compare NPO with direct fine-tuning methods. Direct fine-tuning baselines achieve only limited reward improvements, while dramatically degrading designability. In particular, Flow-GRPO reduces designability to zero, meaning that none of the generated backbones are designable. 
In contrast, NPO substantially improves the reward while retaining much higher designability. Overall, NPO achieves a better reward–sample quality trade-off than direct fine-tuning.

\par\medskip
\noindent
\begin{minipage}{\linewidth}
    \begin{minipage}[t]{0.47\linewidth}
        \vspace{0pt}
        \textbf{NPO amortizes the cost of test-time noise optimization.}
        \Cref{tab:bio_test_time} compares NPO with test-time optimization methods. NPO achieves a higher target reward while generating samples approximately \(118\)–\(151\times\) faster than the test-time baselines. These results show that NPO effectively amortizes test-time search into training, achieving higher reward at substantially lower inference cost.
    \end{minipage}\hfill
    \begin{minipage}[t]{0.50\linewidth}
        \vspace{0pt}
        \centering
        \resizebox{\linewidth}{!}{%

\begin{tabular}{lrrrr}
\toprule
Method & SS Composition & NFE & Time (s) & Rel. cost \\
\midrule
SiD-Protein (Base) & $-0.85 \pm 0.01$ & 5     & 0.07  & $1.0\times$ \\
BoN-256            & $-0.29 \pm 0.02$ & 1,280 & 17.71 & $259.6\times$ \\
ZeNO               & $-0.73 \pm 0.01$ & 2,410 & 22.72 & $333.1\times$ \\
\midrule
\textbf{NPO (Ours)} & $0.81 \pm 0.24$ & 5 & 0.15 & $2.2\times$ \\
\bottomrule
\end{tabular}%
        }
        \captionof{table}{Comparison of NPO with test-time search on biological structure generation tasks. NPO achieves the highest composition score with five evaluations.}
        \label{tab:bio_test_time}
    \end{minipage}
\end{minipage}
\par\medskip

\textbf{Efficiency of NPO compared to noise-policy baselines.} Panel (c) of \Cref{fig:bio_figure} compares NPO with noise-policy learning baselines. NPO achieves higher target reward under the same reward-evaluation budget, consistently demonstrating that ELBO-based training improves sample efficiency.

\section{Conclusion}
\label{sec:conclusion}

We study reward alignment of few-step generative models without access to reward gradients. We introduce Noise Policy Optimization (NPO), which keeps the pretrained generator frozen and learns an amortized flow policy over its input noise to sample from a reward-tilted noise distribution. NPO further uses ELBO-based likelihood estimation to avoid dependence on trajectory-based objectives. 
Extensive experiments across image, language, and protein generation confirm that NPO provides an effective way to adapt few-step generators while preserving their sample quality and fast inference.

\textbf{Limitation and future works.} Although the noise policy is lightweight, it still requires about $2.2$-$2.6\times$ the inference cost of the frozen generator. Distilling the noise policy into a one-step or few-step map could further reduce this overhead to nearly match the speed of the frozen generator. 

Moreover, NPO targets $p^*(x)\propto p_{\mathrm{ref}}(x)\exp(r(x)/\tau)$, the reward-tilted distribution that is commonly used in reference-regularized generative-model alignment. This target is indeed only supported on the reference distribution. However, this support restriction is shared by all methods targeting this distribution, including many of our baselines, rather than being unique to NPO. 

Discovering high-reward samples outside the frozen generator’s support is an important direction for future research. 
Nevertheless, we expect noise-space alignment to become increasingly attractive as pretrained generators improve. As larger and more diverse pretraining datasets bring more high-reward samples within their generative capabilities, NPO can effectively leverage this knowledge, remaining a computationally efficient and effective alternative to direct fine-tuning that scales with pretrained models.

\ificlrfinal
\section*{Acknowledgments}
We thank Jaemoo Choi for valuable discussions and insights that helped shape the direction of our proposed method.
This work was supported by the National Research Foundation of Korea (NRF) grants funded by the Korea government (MSIT) (No.~RS-2022-NR068758 and No.~RS-2024-00410082).
We thank Amazon Web Services (AWS) for sponsoring part of the computational resources used in this work.
Some of the computing for this project was performed on the Sherlock cluster. We would like to thank Stanford University and Stanford Research Computing for providing computational resources and support that contributed to these research results.
\fi

\section*{AI use statement}
We use large language models (LLMs) only for two auxiliary tasks: (1) minor polishing
of the manuscript text for improving grammar and readability, and (2) limited assistance in code
implementation for debugging syntax or refactoring functions. 
LLMs did not contribute
to the research problem formulation, core methodology, or the design of experiments. 
All critical ideas, methods, and analyses presented in this paper are the original work of the authors.

\bibliography{main}
\bibliographystyle{iclr2027_conference}

\clearpage
\appendix
\crefalias{section}{appendix}
\crefalias{subsection}{appendix}
\numberwithin{equation}{section}
\numberwithin{proposition}{section}
\renewcommand{\theHequation}{app.\thesection.\arabic{equation}}
\renewcommand{\theHproposition}{app.\thesection.\arabic{proposition}}
\clearpage
\section{Reward-Tilted Target Distributions}
\label{app:targets}

\subsection{Solution to KL-Regularized Reward Maximization}
\label{app:kl-solution}

Let $\mu$ be a reference distribution, $F(y)$ a reward function, and $\alpha>0$. Define $Z_F=\mathbb E_{y\sim\mu}[\exp(F(y)/\alpha)]$, and assume that $0<Z_F<\infty$ and that the required reward expectations and KL divergences are finite. For
\begin{equation*}
    J(\nu)
    =
    \mathbb E_{y\sim\nu}[F(y)]
    -
    \alpha D_{\mathrm{KL}}(\nu\Vert\mu),
\end{equation*}
define the reward-tilted distribution $\nu^\star(y)=\mu(y)\exp(F(y)/\alpha)/Z_F$. Substituting its definition gives
\begin{equation}
    \label{eq:a-kl-identity}
    J(\nu)
    =
    \alpha\log Z_F
    -
    \alpha D_{\mathrm{KL}}(\nu\Vert\nu^\star).
\end{equation}
Hence, maximizing $J(\nu)$ is equivalent to minimizing $D_{\mathrm{KL}}(\nu\Vert\nu^\star)$, and the optimal solution is $\nu = \nu^\star$. We apply this result to the output and noise spaces as follows.

\textbf{Output-space target.} Setting $y=x$, $\mu=p_{\mathrm{base}}$, and $F(x)=r(x)$ yields the reward-tilted output distribution $p^\star$ in \Cref{eq:tilted-output}, with $Z_F=Z$.

\textbf{Noise-space target.} Setting $y=z$, $\mu=q_{\mathrm{base}}$, and $F(z)=r(g_\theta(z))$ yields the reward-tilted noise distribution $q^\star$ in \Cref{eq:tilted-noise}, with the same normalizer $Z$.

\subsection{Proof of the Noise-Space Pushforward Identity}
\label{app:proof-pushforward}

We show that $g_{\theta\#}q^\star=p^\star$. For any bounded measurable function $f$,
\begin{equation}
    \label{eq:a-pushforward}
    \begin{aligned}
    \mathbb E_{z\sim q^\star}[f(g_\theta(z))]
    &=
    \frac{1}{Z}
    \mathbb E_{z\sim q_{\mathrm{base}}}
    \left[
    f(g_\theta(z))
    e^{r(g_\theta(z))/\alpha}
    \right]
    \\
    &=
    \frac{1}{Z}
    \mathbb E_{x\sim p_{\mathrm{base}}}
    \left[
    f(x)e^{r(x)/\alpha}
    \right]
    =
    \mathbb E_{x\sim p^\star}[f(x)].
    \end{aligned}
\end{equation}
The second equality follows from $p_{\mathrm{base}}=g_{\theta\#}q_{\mathrm{base}}$. Since the equality holds for every bounded measurable $f$, we obtain $g_{\theta\#}q^\star=p^\star$.

\clearpage
\section{Implementation Details of NPO}
\label{app:policy-optimization}

This section describes the implementation details of NPO: the reference flow (\Cref{app:reference-field}), the reference regularization (\Cref{app:reference-regularization}), the estimation of the ELBO-based likelihood (\Cref{app:elbo-weighting}), and the architecture and training of the noise policy (\Cref{app:policy-architecture}).

\subsection{Reference Flow}
\label{app:reference-field}

To construct the Gaussian reference flow, let $\xi,\epsilon\overset{\mathrm{i.i.d.}}{\sim}\mathcal N(0,I)$ and consider $z_t=a_t\xi+\sigma_t\epsilon$. Its conditional velocity is $\dot a_t\xi+\dot\sigma_t\epsilon$, and taking the conditional expectation given $z_t$ gives
\begin{equation}
    \label{eq:b-reference}
    v_{\mathrm{ref}}(z_t,t)
    =
    \frac{a_t\dot a_t+\sigma_t\dot\sigma_t}{a_t^2+\sigma_t^2}\,z_t ,
\end{equation}
which has $q_{\mathrm{base}}=\mathcal N(0,I)$ at both $t=0$ and $t=1$. We use the linear interpolation $a_t=1-t$ and $\sigma_t=t$, which gives $v_{\mathrm{ref}}(z_t,t)=\frac{2t-1}{(1-t)^2+t^2}z_t$. The variance-preserving interpolation $a_t=\cos(\pi t/2)$ and $\sigma_t=\sin(\pi t/2)$, which gives $v_{\mathrm{ref}}=0$, is compared in \Cref{app:ablation}.

\subsection{Reference Regularization}
\label{app:reference-regularization}

The KL divergence in \Cref{eq:noise-objective} can be approximated through the Girsanov theorem as a weighted velocity difference between the policy and the reference flow,
\begin{equation}
    \label{eq:b-ref-penalty}
    D_{\mathrm{KL}}(q_\psi\Vert q_{\mathrm{base}})
    \approx
    \mathbb E_{t}\mathbb E_{z_t}
    \left[
    w_{\mathrm{KL}}(t)
    \left\Vert v_\psi(z_t,t)-v_{\mathrm{ref}}(z_t,t)\right\Vert_2^2
    \right],
\end{equation}
where the weight $w_{\mathrm{KL}}(t)$ is determined by the interpolation schedule. As in prior work~\citep{choi2026rethinking,zheng2026diffusionnft,xue2025advantage}, we use the uniform approximation $w_{\mathrm{KL}}(t)=1$. We additionally scale this penalty by a constant coefficient $\lambda_{\mathrm{ref}}>0$, which gives the reference term $\lambda_{\mathrm{ref}}\Vert v_\psi(z_t,t)-v_{\mathrm{ref}}(z_t,t)\Vert_2^2$ in \Cref{eq:npo-loss}. The value of $\lambda_{\mathrm{ref}}$ is chosen separately for each experiment and is specified in \Cref{app:baselines,app:lang-details,app:bio-details}. Other values of $\lambda_{\mathrm{ref}}$ are compared in \Cref{app:ablation}.

\subsection{ELBO Weighting and Estimation}
\label{app:elbo-weighting}

The ELBO-based likelihood estimator in \Cref{eq:elbo-likelihood} uses the regression score $\ell_\psi(z_0;t,\epsilon)=w(t)\Vert v_\psi(z_t,t)-u_t\Vert_2^2$, where $z_t=a_tz_0+\sigma_t\epsilon$ and $u_t=\dot a_tz_0+\dot\sigma_t\epsilon$ with $t\sim\mathcal U(0,1)$ and $\epsilon\sim\mathcal N(0,I)$. Estimators differ mainly in the weighting $w(t)$, and following \citet{choi2026rethinking} we consider three choices. The path-KL weighting $w(t)=(1-t)/t$ follows from the original derivation of score-based diffusion models, which simplifies the KL divergence between the forward and reverse path measures \citep{song2021score}. For the linear interpolation used here, the path-KL weighting $w(t)=(1-t)/t$ connects the velocity-regression loss to the negative ELBO of the corresponding diffusion model~\citep{song2021maximum,kingma2023understanding}. The simple weighting $w(t)=1$ is the constant weighting used in standard diffusion training \citep{ho2020denoising}. The adaptive weighting self-normalizes the objective for numerical robustness \citep{yin2024one,zheng2026diffusionnft},
\begin{equation}
    \label{eq:b-adaptive}
    \ell_\psi^{\mathrm{adaptive}}
    =
    \frac{t\cdot d\cdot\Vert v_\psi(z_t,t)-u_t\Vert_2^2}
    {\operatorname{sg}\!\left(\Vert v_\psi(z_t,t)-u_t\Vert_1\right)},
\end{equation}
where $d$ is the dimension of $z_0$. We use the adaptive weighting in all experiments. The three weightings are compared in \Cref{app:ablation}.

\textbf{Monte Carlo estimation.} The expectation over $(t,\epsilon)$ in \Cref{eq:elbo-likelihood} is estimated with $M$ independently sampled forward-noising pairs per endpoint,
\begin{equation}
    \label{eq:b-mc}
    \mathbb E_{t,\epsilon}\left[\ell_\psi(z_0;t,\epsilon)\right]
    \approx
    \frac{1}{M}\sum_{j=1}^{M}\ell_\psi(z_0;t_j,\epsilon_j).
\end{equation}
We use $M=3$ in all experiments. Other values of $M$ are compared in \Cref{app:ablation}.

\subsection{Noise Policy Architecture and Training}\label{app:policy-architecture}

\Cref{tab:npo-shared} lists the architecture and optimization settings of the noise policy. 
For images, we use a conditional 2D U-Net from \texttt{diffusers}, conditioned on the prompt embedding of the generator's text encoder through cross-attention. For language and protein backbones, the noise is a one-dimensional sequence without a spatial grid, so we use a DiT~\citep{peebles2023scalable} whose tokens are the positions of the generator input. The flow time $t$ enters through a sinusoidal embedding in both architectures. Within each domain, NPO and the noise-policy baselines, Noise Flow-GRPO and ODS, use the same backbone. We draw endpoints $z_0\sim q_\psi$ by integrating the policy flow from $z_1\sim q_{\mathrm{base}}$ with a Heun solver~\citep{karras2022elucidating}. We train the policy with AdamW~\citep{loshchilov2018decoupled} and set the EMA decay in \Cref{alg:npo} to $\gamma_k$ at round $k$. The learning rate and $\lambda_{\mathrm{ref}}$ are specified per domain in \Cref{app:baselines,app:lang-details,app:bio-details}, and the batch size in \Cref{app:baseline-impl}.

\begin{table}[h]
\centering
\caption{Architecture and optimization settings of the noise policy.}
\label{tab:npo-shared}
\small
\begin{tabular}{@{}l ccc@{}}
\toprule
\multicolumn{4}{@{}l}{\textit{Domain-specific}} \\
\midrule
& Image & Language & Protein \\
\cmidrule(l){2-4}
Backbone & U-Net & DiT & DiT \\
Channel widths / hidden size & $(48,96,192)$ & $256$ & $256$ \\
Layers & $2$ per resolution & depth $6$, $4$ heads & depth $6$, $4$ heads \\
Weight decay & $10^{-4}$ & $10^{-4}$ & $0$ \\
\midrule
\multicolumn{4}{@{}l}{\textit{Shared across domains}} \\
\midrule
Policy solver & \multicolumn{3}{c}{Heun, $10$ steps} \\
Optimizer & \multicolumn{3}{c}{AdamW, $(\beta_1,\beta_2)=(0.9,0.999)$} \\
Gradient-norm clipping & \multicolumn{3}{c}{$5.0$} \\
Gradient updates per round & \multicolumn{3}{c}{$1$} \\
EMA decay $\gamma_k$ & \multicolumn{3}{c}{$\min(0.001\,k,\,0.5)$} \\
Interpolation / ELBO weighting / $M$ & \multicolumn{3}{c}{linear / adaptive / $3$} \\
\bottomrule
\end{tabular}
\end{table}

\clearpage
\section{Implementation of the Baselines}
\label{app:baseline-impl}

We reimplement all baselines in our experimental framework, using their official implementations as references when available. For ZeNO, whose official implementation is not publicly available, we follow the algorithm of \citet{kim2026gradient}. Within each domain, all methods share the generator, the reward, and the evaluation protocol. Noise Flow-GRPO and ODS use the noise-policy backbone of NPO (\Cref{tab:npo-shared}). Flow-GRPO and FAV-B fine-tune the attention layers of the generator with LoRA~\citep{hu2022lora}. We set the LoRA scale to $\alpha=2r$ and choose the rank $r$ so that the number of trainable parameters matches that of the noise policy (\Cref{tab:baseline-lora}). All methods except FAV-B draw $64$ samples per training step. For images, NPO, Noise Flow-GRPO, and Flow-GRPO use $B=8$ prompts with a group of $G=8$ samples each, and ODS uses $64$ prompts with one sample each. The language and protein generators are unconditional, and all methods draw $64$ samples per step.

\begin{table}[h]
\centering
\caption{LoRA rank of the direct baselines and the resulting number of trainable parameters.}
\label{tab:baseline-lora}
\small
\begin{tabular}{@{}l c c c@{}}
\toprule
Generator & LoRA rank $r$ & LoRA parameters & Noise-policy parameters \\
\midrule
SANA-Sprint & $28$  & $14.45$M & $14.43$M \\
SD-Turbo    & $68$  & $14.11$M & $14.16$M \\
ELF         & $137$ & $7.58$M  & $7.58$M \\
Protein     & $106$ & $7.33$M  & $7.32$M \\
\bottomrule
\end{tabular}
\end{table}

\paragraph{Noise Flow-GRPO.} Noise Flow-GRPO trains the noise policy with the GRPO objective of \citet{liu2025flowgrpo}, with the reference flow of \Cref{app:reference-field} as the reference model. During training, it samples the policy flow with a $10$-step Euler--Maruyama discretization of an SDE with noise level $a=0.7$, and it clips the importance ratio at $\epsilon=1\text{e-}4$. At evaluation, it samples the policy flow with a $10$-step Heun solver. We train with AdamW, $(\beta_1,\beta_2)=(0.9,0.999)$, weight decay $10^{-4}$ ($0$ for proteins), and gradient-norm clipping at $1.0$ ($5.0$ for proteins).

\paragraph{Flow-GRPO.} Flow-GRPO fine-tunes the generator with the GRPO objective of \citet{liu2025flowgrpo}, with the frozen generator as the reference model. To obtain transition likelihoods, we add Gaussian noise after each sampling step for multi-step sampling and to the output for one-step sampling. We clip the importance ratio at $\epsilon=0.2$ for images and $\epsilon=1\text{e-}4$ otherwise. At evaluation, the generator samples without the added noise. We train with AdamW, $(\beta_1,\beta_2)=(0.9,0.999)$, weight decay $10^{-4}$, and gradient-norm clipping at $1.0$.

\paragraph{FAV-B.} We follow the black-box variant of \citet{lee2026aligning}. For each of $C$ conditions, FAV-B draws $K$ particles from the fine-tuned generator and $K$ particles from the frozen generator. It estimates the reward gradient at each particle with an antithetic evolution-strategies estimate over $N$ random perturbations of scale $\sigma$, which costs $K(2N+1)$ reward calls per condition. It adds $\beta$ times this gradient to the kernel density estimate of the prior score and regresses the LoRA onto the particles moved by the resulting velocity. We use $K=8$ and $N=4$, with $C=4$ and $\sigma=0.1$ for images, $C=1$ and $\sigma=0.1$ for language, and $C=1$, $N=16$, and $\sigma$ equal to $0.01$ times the particle scale for proteins. The kernel bandwidths follow the median heuristic. We train with AdamW, $(\beta_1,\beta_2)=(0.9,0.95)$, weight decay $0.01$, and gradient-norm clipping at $2.0$.

\paragraph{ODS.} We follow \citet{venkatraman2025outsourced} and use the noise-policy backbone of NPO as the score network. ODS samples the noise with a $25$-step Euler--Maruyama discretization of a cosine variance-preserving SDE, during training and at evaluation. It trains with the relative trajectory balance objective toward $q_{\mathrm{base}}(z)\,r(g_\theta(z))^{\beta}$ and sets $\log Z$ to the batch mean. Each step draws trajectories from the prior with probability $0.2$, from a uniform replay buffer of size $10{,}000$ with probability $0.2$, and from the current policy otherwise. We use $\log(r+2)$ for compressibility and the exponential of the secondary-structure diversity score for proteins so that $\log r$ is defined. We train with Adam and gradient-norm clipping at $0.1$.

\paragraph{Test-time baselines.} Best-of-$N$ (BoN), ReNO~\citep{eyring2024reno}, and ZeNO~\citep{kim2026gradient} search the initial noise of each sample at inference time and leave the generator unchanged. For each method, we search as many samples as the evaluation protocol of each domain uses and evaluate them in the same way as the other methods.
\begin{itemize}
    \item \textbf{BoN} draws $N$ initial noises for each sample and keeps the one with the highest reward.
    \item \textbf{ReNO} optimizes the initial noise by gradient ascent on the reward, which requires a differentiable reward, so we run it only on SD-Turbo with one sampling step under the aesthetic reward. We use the default hyperparameters of \citet{eyring2024reno}: SGD with Nesterov momentum $0.9$, learning rate $5.0$, gradient-norm clipping at $0.1$, and a norm regularization weight of $0.01$.
    \item \textbf{ZeNO} optimizes the initial noise with a zeroth-order estimate of the reward gradient. We use the default hyperparameters of \citet{kim2026gradient}, $\beta=0.01$, $\eta=1.5$, and $N=16$ perturbations per iteration, in all domains.
\end{itemize}

\clearpage
\section{Image Experiment Details}
\label{app:baselines}

\subsection{Generator, Rewards and Evaluation Metrics}
\label{app:image-rewards}

\paragraph{Generator.} We use two frozen text-to-image generators. SANA-Sprint~\citep{chen2025sana} is a distilled $0.6$B-parameter model that generates $1024\times1024$ images from a $32\times32\times32$ latent. We run it at one and four sampling steps with guidance scale $4.5$. SD-Turbo~\citep{sauer2024adversarial} is an adversarially distilled Stable Diffusion 2.1 model that generates $512\times512$ images from a $4\times64\times64$ latent. We run it at one and four sampling steps.

\paragraph{Target rewards.} We use two target rewards. The LAION aesthetic predictor~\citep{schuhmann2022laion} is a linear head on a frozen CLIP image embedding~\citep{radford2021learning} and maps an image to a scalar aesthetic score. JPEG compressibility~\citep{black2024training} is the negative file size of the image after JPEG encoding.

\paragraph{Evaluation metrics.} We measure semantic collapse with ImageReward~\citep{xu2023imagereward}, a learned reward model of prompt--image alignment that no method optimizes. We measure diversity collapse with LPIPS~\citep{zhang2018unreasonable}, the perceptual distance between two images, averaged over all pairs of samples that share a prompt.

\subsection{Hyperparameters}
\label{app:baselines-hp}
The learning rate and the coefficient of each method are swept over the ranges in \Cref{tab:image-hp-grid}. For hyperparameter selection, we excluded configurations where the target reward dropped abruptly or consistently decreased during training. Among the remaining configurations, we select the best setting whose target reward rises stably across 3 different seeds. We use the same strategy to select this hyperparameter for all methods, including our NPO. We use 3 seeds to select the hyperparameters and run two additional seeds for the selected ones. \Cref{tab:image-hp-chosen,tab:sdturbo-hp-chosen} list the selected settings.

\begin{table}[h]
\centering
\caption{Swept ranges in the image domain. $\lambda_{\mathrm{ref}}$ is the reference coefficient of NPO. $\beta$ is the KL coefficient of Noise Flow-GRPO and Flow-GRPO, the reward exponent of ODS, and the reward-gradient scale of FAV-B.}
\label{tab:image-hp-grid}
\small
\begin{tabular}{@{}l l@{}}
\toprule
Method & Coefficient \\
\midrule
\multicolumn{2}{@{}l}{\textit{SANA-Sprint}, learning rate $\in \{3\text{e-}5, 1\text{e-}4, 3\text{e-}4, 1\text{e-}3, 3\text{e-}3\}$} \\
\midrule
NPO (ours)         & $\lambda_{\mathrm{ref}} \in \{0, 3\text{e-}5, 1\text{e-}4, 3\text{e-}4, 1\text{e-}3\}$ \\
Noise Flow-GRPO    & $\beta \in \{0, 4\text{e-}3, 1\text{e-}2, 4\text{e-}2, 1\text{e-}1\}$ \\
ODS                & $\beta \in \{50, 100, 500, 1000, 5000\}$ \\
Flow-GRPO & $\beta \in \{0, 1\text{e-}3, 1\text{e-}2, 1\text{e-}1, 1\}$ \\
FAV-B              & $\beta \in \{1\text{e-}3, 1\text{e-}2, 1\text{e-}1, 1\}$ \\
\midrule
\multicolumn{2}{@{}l}{\textit{SD-Turbo}, learning rate $\in \{1\text{e-}5, 3\text{e-}5, 1\text{e-}4, 3\text{e-}4, 1\text{e-}3, 3\text{e-}3\}$} \\
\midrule
NPO (ours)         & $\lambda_{\mathrm{ref}} \in \{3\text{e-}5, 1\text{e-}4, 3\text{e-}4, 1\text{e-}3, 3\text{e-}3\}$ \\
Noise Flow-GRPO    & $\beta \in \{4\text{e-}3, 1\text{e-}2, 4\text{e-}2, 1\text{e-}1\}$ \\
ODS                & $\beta \in \{50, 100, 500, 1000, 5000\}$ \\
Flow-GRPO & $\beta \in \{1\text{e-}3, 1\text{e-}2, 1\text{e-}1, 1\}$ \\
FAV-B              & $\beta \in \{1\text{e-}3, 1\text{e-}2, 1\text{e-}1, 1\}$ \\
\bottomrule
\end{tabular}
\end{table}

\begin{table}[h]
\centering
\caption{Selected settings on SANA-Sprint.}
\label{tab:image-hp-chosen}
\small
\begin{tabular}{@{}l ll ll@{}}
\toprule
& \multicolumn{2}{c}{Aesthetic} & \multicolumn{2}{c}{Compressibility} \\
\cmidrule(lr){2-3}\cmidrule(lr){4-5}
Method & Learning rate & Coefficient & Learning rate & Coefficient \\
\midrule
NPO (ours)         & $3\text{e-}5$ & $\lambda_{\mathrm{ref}}=1\text{e-}4$ & $3\text{e-}5$ & $\lambda_{\mathrm{ref}}=3\text{e-}4$ \\
Noise Flow-GRPO    & $3\text{e-}5$ & $\beta=1\text{e-}2$  & $3\text{e-}5$ & $\beta=1\text{e-}1$ \\
ODS                & $3\text{e-}5$ & $\beta=500$   & $3\text{e-}5$ & $\beta=500$ \\
Flow-GRPO & $3\text{e-}4$ & $\beta=1\text{e-}3$ & $1\text{e-}3$ & $\beta=1\text{e-}3$ \\
FAV-B              & $3\text{e-}4$ & $\beta=1$     & $1\text{e-}3$ & $\beta=1$ \\
\bottomrule
\end{tabular}
\end{table}

\begin{table}[h]
\centering
\caption{Selected settings on SD-Turbo.}
\label{tab:sdturbo-hp-chosen}
\small
\begin{tabular}{@{}l ll ll@{}}
\toprule
& \multicolumn{2}{c}{Aesthetic} & \multicolumn{2}{c}{Compressibility} \\
\cmidrule(lr){2-3}\cmidrule(lr){4-5}
Method & Learning rate & Coefficient & Learning rate & Coefficient \\
\midrule
NPO (ours)                  & $3\text{e-}5$ & $\lambda_{\mathrm{ref}}=3\text{e-}4$ & $1\text{e-}5$ & $\lambda_{\mathrm{ref}}=1\text{e-}4$ \\
Noise Flow-GRPO             & $3\text{e-}5$ & $\beta=4\text{e-}2$  & $1\text{e-}5$ & $\beta=1\text{e-}1$ \\
ODS                         & $3\text{e-}5$ & $\beta=500$   & $1\text{e-}5$ & $\beta=500$ \\
Flow-GRPO (four steps)      & $3\text{e-}4$ & $\beta=1\text{e-}2$  & $1\text{e-}4$ & $\beta=1\text{e-}3$ \\
Flow-GRPO (one step)        & $3\text{e-}4$ & $\beta=1\text{e-}2$  & $1\text{e-}4$ & $\beta=1\text{e-}3$ \\
FAV-B                       & $1\text{e-}4$ & $\beta=1$     & $1\text{e-}4$ & $\beta=1$ \\
\bottomrule
\end{tabular}
\end{table}

\subsection{Evaluation Protocol}
\label{app:baselines-eval}

We evaluate on $45$ simple-animal prompts~\citep{black2024training} with $16$ samples per prompt. We score every sample with the target reward and ImageReward and report their means. We compute LPIPS between all pairs of samples that share a prompt and report the mean distance. We run each setting with five seeds and report the mean and standard deviation across seeds.

\clearpage
\section{Language Experiment Details}
\label{app:lang-details}

\subsection{Generator, Rewards and Evaluation Metrics}
\label{app:lang-rewards}

\paragraph{Generator.} We use the progressively distilled four-step ELF model~\citep{hu2026elf} trained on OpenWebText. The model is unconditional and generates sequences of $1024$ tokens from a $1024\times512$ continuous latent. We run it with its probability-flow ODE sampler. ELF draws its sampling times at random from a logit-normal distribution. We instead place them at the quantiles of the same distribution, which makes the sampler deterministic given the initial noise.

\paragraph{Target reward.} The target reward is the entailment probability of the DeBERTa-v3-large NLI model~\citep{he2021debertav3} fine-tuned on MNLI and related NLI corpora~\citep{williams2018broad}. The premise is the generated text, and the hypothesis is ``This text is about \textit{topic}.'' for $\textit{topic} \in \{\text{politics}, \text{science}\}$.

\paragraph{Evaluation metrics.} We measure fluency with the perplexity of GPT-2~Large~\citep{radford2019language}, the exponentiated mean negative log-likelihood of a generated sequence. We measure diversity with the unigram entropy~\citep{lee2026flow} $H=-\sum_v p_v\log p_v$ of a sample, where $p_v$ is the frequency of token $v$ in the sample under the GPT-2 tokenizer.

\subsection{Hyperparameters}
\label{app:lang-hp}

The learning rate and the coefficient of each method are swept over the ranges in \Cref{tab:lang-hp-grid}. For hyperparameter selection, we excluded configurations where the target reward dropped abruptly or consistently decreased during training. Among the remaining configurations, we select the best setting whose target reward rises stably across 3 different seeds. We use the same strategy to select this hyperparameter for all methods, including our NPO. We use 3 seeds to select the hyperparameters and run two additional seeds for the selected ones. \Cref{tab:lang-hp-chosen} lists the selected settings.

\begin{table}[h]
\centering
\caption{Swept ranges on the topic-steering task. $\lambda_{\mathrm{ref}}$ is the reference coefficient of NPO. $\beta$ is the KL coefficient of Noise Flow-GRPO and Flow-GRPO, the reward exponent of ODS, and the reward-gradient scale of FAV-B.}
\label{tab:lang-hp-grid}
\small
\begin{tabular}{@{}l l@{}}
\toprule
Method & Coefficient \\
\midrule
\multicolumn{2}{@{}l}{\textit{All methods}, learning rate $\in \{1\text{e-}4, 3\text{e-}4, 1\text{e-}3, 3\text{e-}3\}$} \\
\midrule
NPO (ours)         & $\lambda_{\mathrm{ref}} \in \{0, 3\text{e-}5, 1\text{e-}4, 3\text{e-}4, 1\text{e-}3\}$ \\
Noise Flow-GRPO    & $\beta \in \{0, 4\text{e-}3, 1\text{e-}2, 4\text{e-}2, 1\text{e-}1\}$ \\
ODS                & $\beta \in \{50, 100, 500, 1000, 5000\}$ \\
Flow-GRPO & $\beta \in \{0, 1\text{e-}3, 1\text{e-}2, 1\text{e-}1, 1\}$ \\
FAV-B              & $\beta \in \{1\text{e-}3, 1\text{e-}2, 1\text{e-}1, 1\}$ \\
\bottomrule
\end{tabular}
\end{table}

\begin{table}[h]
\centering
\caption{Selected settings on the topic-steering task.}
\label{tab:lang-hp-chosen}
\small
\begin{tabular}{@{}l ll ll@{}}
\toprule
& \multicolumn{2}{c}{Politics} & \multicolumn{2}{c}{Science} \\
\cmidrule(lr){2-3}\cmidrule(lr){4-5}
Method & Learning rate & Coefficient & Learning rate & Coefficient \\
\midrule
NPO (ours)         & $1\text{e-}4$ & $\lambda_{\mathrm{ref}}=3\text{e-}4$ & $1\text{e-}4$ & $\lambda_{\mathrm{ref}}=3\text{e-}4$ \\
Noise Flow-GRPO    & $1\text{e-}4$ & $\beta=1\text{e-}2$  & $1\text{e-}4$ & $\beta=1\text{e-}2$ \\
ODS                & $1\text{e-}4$ & $\beta=50$    & $1\text{e-}4$ & $\beta=50$ \\
Flow-GRPO & $1\text{e-}3$ & $\beta=1\text{e-}3$ & $1\text{e-}3$ & $\beta=1\text{e-}2$ \\
FAV-B              & $1\text{e-}3$ & $\beta=1$     & $1\text{e-}3$ & $\beta=1$ \\
\bottomrule
\end{tabular}
\end{table}

\subsection{Evaluation Protocol}
\label{app:lang-eval}

We evaluate on $256$ samples. We score every sample with the target reward, perplexity, and unigram entropy and report their means. We run each setting with five seeds and report the mean and standard deviation across seeds.

\clearpage
\section{Biological Experiment Details}
\label{app:bio-details}

\subsection{Generator, Reward and Evaluation Metrics}
\label{app:bio-rewards}

\paragraph{Generator.} We use the five-step protein backbone generator of \citet{xie2025distilled}, distilled from Proteina~\citep{geffner2025proteina} with score identity distillation. We run it at $K=5$ sampling steps and generate backbones of $L=150$ residues. We sample without re-injecting noise between steps, which makes the sampler deterministic given the initial noise.

\paragraph{Target reward.} The target reward is the secondary-structure diversity score of \citet{venkatraman2025outsourced}, computed from the proportions of $\alpha$-helix, $\beta$-sheet, and coil residues in a backbone. We assign the secondary structure with P-SEA~\citep{labesse1997psea} from the C$\alpha$ trace instead of DSSP, which requires backbone atoms that the generator does not produce.

\paragraph{Evaluation metrics.} We measure designability with the self-consistency protocol of \citet{geffner2025proteina} and \citet{xie2025distilled}. ProteinMPNN~\citep{dauparas2022robust} designs eight sequences for each backbone, and ESMFold~\citep{lin2023evolutionary} folds each sequence. A backbone is designable when the smallest C$\alpha$ RMSD between the backbone and a refolded structure is below $2$\,\AA{}, and we report the designable fraction. We measure diversity with the mean TM-score~\citep{zhang2004scoring} over all pairs of backbones, computed with TM-align~\citep{zhang2005tm}. A lower value indicates higher diversity.

\subsection{Hyperparameters}
\label{app:bio-hp}

The learning rate and the coefficient of each method are swept over the ranges in \Cref{tab:bio-hp-grid}. For hyperparameter selection, we excluded configurations where the target reward dropped abruptly or consistently decreased during training. Among the remaining configurations, we select the best setting whose target reward rises stably across 3 different seeds. We use the same strategy to select this hyperparameter for all methods, including our NPO. We use 3 seeds to select the hyperparameters and run two additional seeds for the selected ones.  \Cref{tab:bio-hp-chosen} lists the selected settings.

\begin{table}[h]
\centering
\caption{Swept ranges in the biological domain. $\lambda_{\mathrm{ref}}$ is the reference coefficient of NPO. $\beta$ is the KL coefficient of Noise Flow-GRPO and Flow-GRPO, the reward exponent of ODS, and the reward-gradient scale of FAV-B.}
\label{tab:bio-hp-grid}
\small
\begin{tabular}{@{}l l@{}}
\toprule
Method & Coefficient \\
\midrule
\multicolumn{2}{@{}l}{\textit{All methods}, learning rate $\in \{1\text{e-}4, 3\text{e-}4, 1\text{e-}3, 3\text{e-}3\}$} \\
\midrule
NPO (ours)         & $\lambda_{\mathrm{ref}} \in \{ 3\text{e-}5, 1\text{e-}4, 3\text{e-}4, 1\text{e-}3\}$ \\
Noise Flow-GRPO     & $\beta \in \{1\text{e-}2, 4\text{e-}2, 1\text{e-}1, 4\text{e-}1\}$ \\
ODS                & $\beta \in \{100, 300, 500, 1000\}$ \\
Flow-GRPO & $\beta \in \{4\text{e-}4, 1\text{e-}3, 4\text{e-}3, 1\text{e-}2\}$ \\
FAV-B              & $\beta \in \{1\text{e-}4, 3\text{e-}4, 1\text{e-}3, 3\text{e-}3, 1\text{e-}2\}$ \\
\bottomrule
\end{tabular}
\end{table}

\begin{table}[h]
\centering
\caption{Selected settings at $L=150$.}
\label{tab:bio-hp-chosen}
\small
\begin{tabular}{@{}l l l@{}}
\toprule
Method & Learning rate & Coefficient \\
\midrule
NPO (ours)         & $3\text{e-}4$ & $\lambda_{\mathrm{ref}} = 1\text{e-}4$ \\
Noise Flow-GRPO     & $3\text{e-}4$ & $\beta = 1\text{e-}2$ \\
ODS                & $3\text{e-}4$ & $\beta = 500$ \\
Flow-GRPO & $1\text{e-}4$ & $\beta = 4\text{e-}4$ \\
FAV-B              & $1\text{e-}4$ & $\beta = 3\text{e-}4$ \\
\bottomrule
\end{tabular}
\end{table}

\subsection{Evaluation Protocol}
\label{app:bio-eval}

We evaluate on $256$ backbones. We compute the target reward on all $256$ backbones and designability and pairwise TM-score on the first $32$. We run each setting with five seeds and report the mean and standard deviation across seeds.

\section{Numerical Results of the Main Experiments}
\label{app:numerical}

We report the values plotted in the figures of the main text. Each entry is the mean $\pm$ standard deviation over five seeds.

\begin{table}[H]
\centering
\small
\caption{\textbf{Numerical results of \Cref{fig:image_target_1step}.} Comparison of NPO and direct fine-tuning baselines with one sampling step. The values are taken at the final checkpoint of each learning curve.}
\label{tab:num-image_target_1step}
\resizebox{\ifdim\width>\linewidth\linewidth\else\width\fi}{!}{%
\begin{tabular}{lcccc}
\toprule
& \multicolumn{2}{c}{Aesthetic reward} & \multicolumn{2}{c}{Compressibility reward} \\
\cmidrule(lr){2-3}\cmidrule(lr){4-5}
Method & SANA-Sprint & SD-Turbo & SANA-Sprint & SD-Turbo \\
\midrule
Pretrained (Base) & $6.28 \pm 0.01$ & $6.11 \pm 0.01$ & $-0.485 \pm 0.002$ & $-0.208 \pm 0.001$ \\
\midrule
FAV-B & $6.28 \pm 0.13$ & $5.97 \pm 0.31$ & $-0.292 \pm 0.011$ & $-0.177 \pm 0.006$ \\
Flow-GRPO & $6.27 \pm 0.02$ & $6.11 \pm 0.02$ & $-0.446 \pm 0.014$ & $-0.203 \pm 0.002$ \\
\textbf{NPO (Ours)} & $6.55 \pm 0.05$ & $6.47 \pm 0.05$ & $-0.134 \pm 0.008$ & $-0.096 \pm 0.003$ \\
\bottomrule
\end{tabular}}
\end{table}

\begin{table}[H]
\centering
\small
\caption{\textbf{Numerical results of \Cref{fig:image_pareto_4step}.} Comparison of NPO and direct fine-tuning baselines on SANA-Sprint with four sampling steps.}
\label{tab:num-image_pareto_4step}
\resizebox{\ifdim\width>\linewidth\linewidth\else\width\fi}{!}{%
\begin{tabular}{lcccccc}
\toprule
& \multicolumn{3}{c}{Aesthetic reward} & \multicolumn{3}{c}{Compressibility reward} \\
\cmidrule(lr){2-4}\cmidrule(lr){5-7}
Method & Aesthetic & ImageReward & LPIPS & Compressibility & ImageReward & LPIPS \\
\midrule
SANA-Sprint (Base) & $6.18 \pm 0.00$ & $0.98 \pm 0.01$ & $0.63 \pm 0.00$ & $-0.515 \pm 0.001$ & $0.98 \pm 0.01$ & $0.64 \pm 0.00$ \\
\midrule
FAV-B & $6.22 \pm 0.08$ & $0.94 \pm 0.05$ & $0.63 \pm 0.05$ & $-0.367 \pm 0.016$ & $0.93 \pm 0.08$ & $0.56 \pm 0.02$ \\
Flow-GRPO & $6.67 \pm 0.18$ & $0.56 \pm 0.39$ & $0.58 \pm 0.05$ & $-0.111 \pm 0.067$ & $-2.00 \pm 0.40$ & $0.11 \pm 0.12$ \\
\textbf{NPO (Ours)} & $6.54 \pm 0.03$ & $1.03 \pm 0.03$ & $0.66 \pm 0.02$ & $-0.185 \pm 0.017$ & $0.32 \pm 0.14$ & $0.32 \pm 0.03$ \\
\bottomrule
\end{tabular}}
\end{table}

\begin{table}[H]
\centering
\small
\caption{\textbf{Numerical results of \Cref{fig:image_noise_efficiency}.} Comparison of NPO and noise-policy learning baselines on few-step generation. The values are taken at the final checkpoint of each learning curve.}
\label{tab:num-image_noise_efficiency}
\resizebox{\ifdim\width>\linewidth\linewidth\else\width\fi}{!}{%
\begin{tabular}{lcccc}
\toprule
& \multicolumn{2}{c}{Aesthetic reward} & \multicolumn{2}{c}{Compressibility reward} \\
\cmidrule(lr){2-3}\cmidrule(lr){4-5}
Method & 1-step & 4-step & 1-step & 4-step \\
\midrule
SANA-Sprint (Base) & $6.29 \pm 0.01$ & $6.18 \pm 0.01$ & $-0.485 \pm 0.002$ & $-0.515 \pm 0.002$ \\
\midrule
ODS & $6.38 \pm 0.01$ & $6.33 \pm 0.04$ & $-0.383 \pm 0.064$ & $-0.329 \pm 0.085$ \\
Noise Flow-GRPO & $6.48 \pm 0.02$ & $6.43 \pm 0.02$ & $-0.208 \pm 0.038$ & $-0.204 \pm 0.033$ \\
\textbf{NPO (Ours)} & $6.55 \pm 0.02$ & $6.54 \pm 0.03$ & $-0.134 \pm 0.008$ & $-0.130 \pm 0.005$ \\
\bottomrule
\end{tabular}}
\end{table}

\begin{table}[H]
\centering
\small
\caption{\textbf{Numerical results of \Cref{fig:language_pareto}.} Comparison of NPO and direct fine-tuning baselines on language generation.}
\label{tab:num-language_pareto}
\resizebox{\ifdim\width>\linewidth\linewidth\else\width\fi}{!}{%
\begin{tabular}{lcccccc}
\toprule
& \multicolumn{3}{c}{Politics} & \multicolumn{3}{c}{Science} \\
\cmidrule(lr){2-4}\cmidrule(lr){5-7}
Method & Reward & Perplexity & Entropy & Reward & Perplexity & Entropy \\
\midrule
ELF (Base) & $0.31 \pm 0.01$ & $42.2 \pm 0.2$ & $5.11 \pm 0.06$ & $0.13 \pm 0.01$ & $42.2 \pm 0.2$ & $5.11 \pm 0.05$ \\
\midrule
FAV-B & $0.35 \pm 0.34$ & $2.2 \pm 1.3$ & $0.26 \pm 0.35$ & $0.47 \pm 0.32$ & $1.5 \pm 0.9$ & $0.18 \pm 0.34$ \\
Flow-GRPO & $0.46 \pm 0.17$ & $2.3 \pm 1.9$ & $0.55 \pm 0.44$ & $0.59 \pm 0.14$ & $2.5 \pm 1.3$ & $0.33 \pm 0.44$ \\
\textbf{NPO (Ours)} & $0.93 \pm 0.02$ & $39.2 \pm 3.0$ & $5.04 \pm 0.04$ & $0.94 \pm 0.01$ & $47.5 \pm 2.7$ & $5.14 \pm 0.05$ \\
\bottomrule
\end{tabular}}
\end{table}

\begin{table}[H]
\centering
\small
\caption{\textbf{Numerical results of \Cref{fig:lang_noise_efficiency}.} Comparison of NPO and noise-policy learning baselines on language generation. The values are taken at the final checkpoint of each learning curve.}
\label{tab:num-lang_noise_efficiency}
\resizebox{\ifdim\width>\linewidth\linewidth\else\width\fi}{!}{%
\begin{tabular}{lcc}
\toprule
Method & Politics & Science \\
\midrule
ELF (Base) & $0.31 \pm 0.01$ & $0.13 \pm 0.01$ \\
\midrule
ODS & $0.58 \pm 0.27$ & $0.46 \pm 0.27$ \\
Noise Flow-GRPO & $0.87 \pm 0.04$ & $0.88 \pm 0.01$ \\
\textbf{NPO (Ours)} & $0.93 \pm 0.02$ & $0.94 \pm 0.01$ \\
\bottomrule
\end{tabular}}
\end{table}

\begin{table}[H]
\centering
\small
\caption{\textbf{Numerical results of \Cref{fig:bio_figure}(a) and (b).} Comparison of NPO and direct fine-tuning baselines on protein structure generation.}
\label{tab:num-bio_figure_ab}
\resizebox{\ifdim\width>\linewidth\linewidth\else\width\fi}{!}{%
\begin{tabular}{lccc}
\toprule
Method & SS Composition & Designability & Pairwise TM \\
\midrule
SiD-Protein (Base) & $-0.86$ & $0.44$ & $0.35$ \\
\midrule
FAV-B & $-0.70 \pm 0.17$ & $0.13 \pm 0.17$ & $0.33 \pm 0.01$ \\
Flow-GRPO & $0.01 \pm 0.39$ & $0.00 \pm 0.00$ & $0.28 \pm 0.02$ \\
\textbf{NPO (Ours)} & $0.81 \pm 0.24$ & $0.30 \pm 0.29$ & $0.57 \pm 0.07$ \\
\bottomrule
\end{tabular}}
\end{table}

\begin{table}[H]
\centering
\small
\caption{\textbf{Numerical results of \Cref{fig:bio_figure}(c).} Comparison of NPO and noise-policy learning baselines on protein structure generation. The values are taken at the final checkpoint of each learning curve.}
\label{tab:num-bio_figure_c}
\resizebox{\ifdim\width>\linewidth\linewidth\else\width\fi}{!}{%
\begin{tabular}{lc}
\toprule
Method & SS Composition \\
\midrule
SiD-Protein (Base) & $-0.86$ \\
\midrule
ODS & $0.41 \pm 0.73$ \\
Noise Flow-GRPO & $-0.37 \pm 0.54$ \\
\textbf{NPO (Ours)} & $0.81 \pm 0.24$ \\
\bottomrule
\end{tabular}}
\end{table}

\clearpage
\section{Additional Results}
\label{app:additional_results}

\subsection{Comparison with Direct Fine-Tuning on SD-Turbo}
\label{app:sdturbo-direct}

\begin{figure}[H]
\centering
\includegraphics[width=\linewidth]{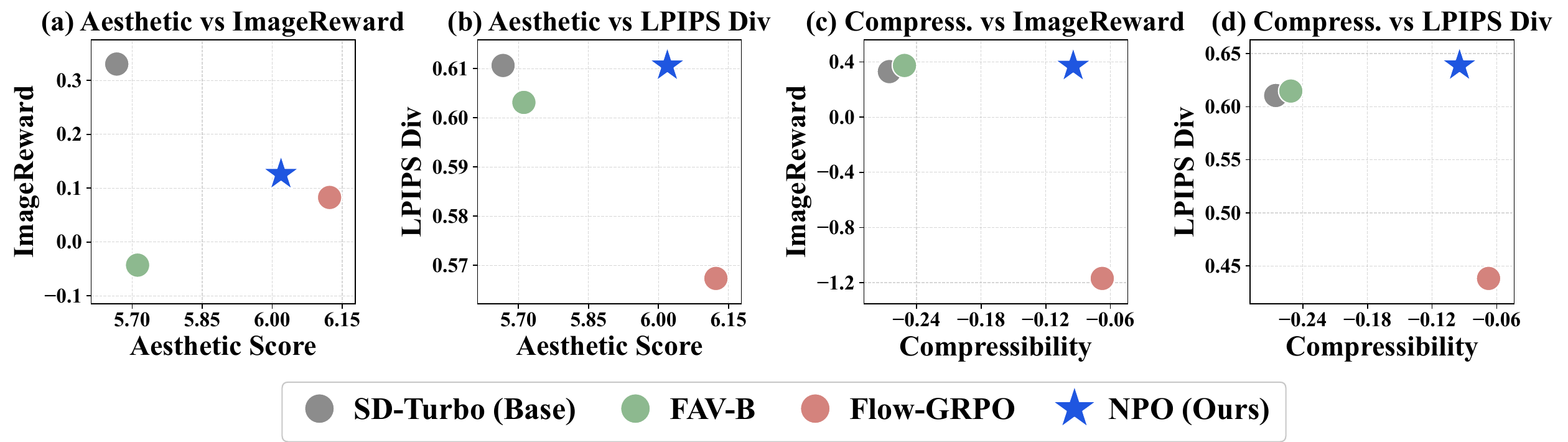}
\caption{\textbf{Reward-quality trade-off of NPO and direct fine-tuning baselines on SD-Turbo with four sampling steps.} (a, b) Aesthetic score against ImageReward and LPIPS diversity. (c, d) Compressibility against ImageReward and LPIPS diversity.}
\label{fig:sdturbo_pareto_4step}
\end{figure}

\Cref{fig:sdturbo_pareto_4step} repeats the four-step comparison of \Cref{fig:image_pareto_4step} on SD-Turbo. FAV-B barely improves either target reward. Flow-GRPO reaches the highest target reward, but its ImageReward and LPIPS diversity decrease substantially, most severely under compressibility. NPO reaches a comparable target reward while keeping LPIPS diversity at or above that of the frozen generator and ImageReward above that of Flow-GRPO.

\subsection{Ablation of the NPO Objective}
\label{app:ablation}

We ablate the design choices of the NPO objective described in \Cref{app:policy-optimization} on SANA-Sprint with one sampling step under the aesthetic reward. Each ablation changes one choice from the default setting and is run with three seeds. NPO is insensitive to most of these choices. Except for the path-KL weighting, every variant performs similarly to the default setting, which shows that NPO is robust to its design choices.

\begin{figure}[H]
\centering
\includegraphics[width=\linewidth]{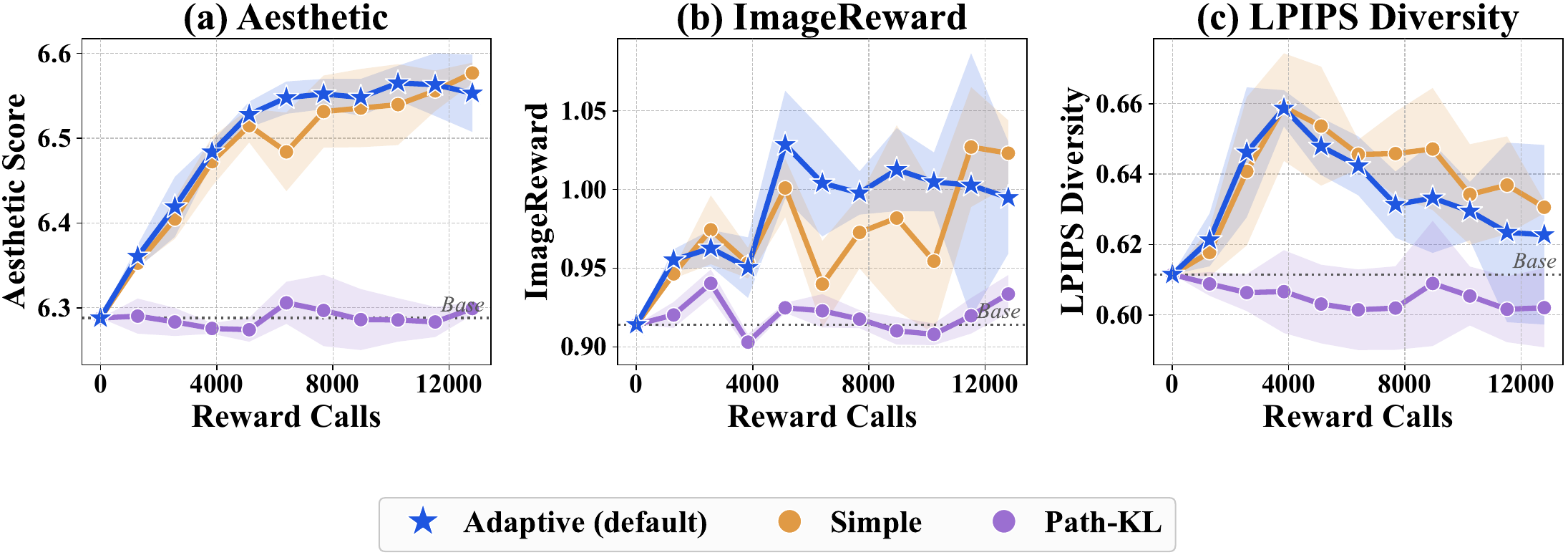}
\caption{\textbf{Ablation of the ELBO weighting.}}
\label{fig:abl-weighting}
\end{figure}

\paragraph{ELBO weighting.} \Cref{fig:abl-weighting} compares the three weightings of \Cref{app:elbo-weighting}. The adaptive and simple weightings perform similarly, whereas the path-KL weighting barely improves the reward. We use the adaptive weighting.

\begin{figure}[H]
\centering
\includegraphics[width=\linewidth]{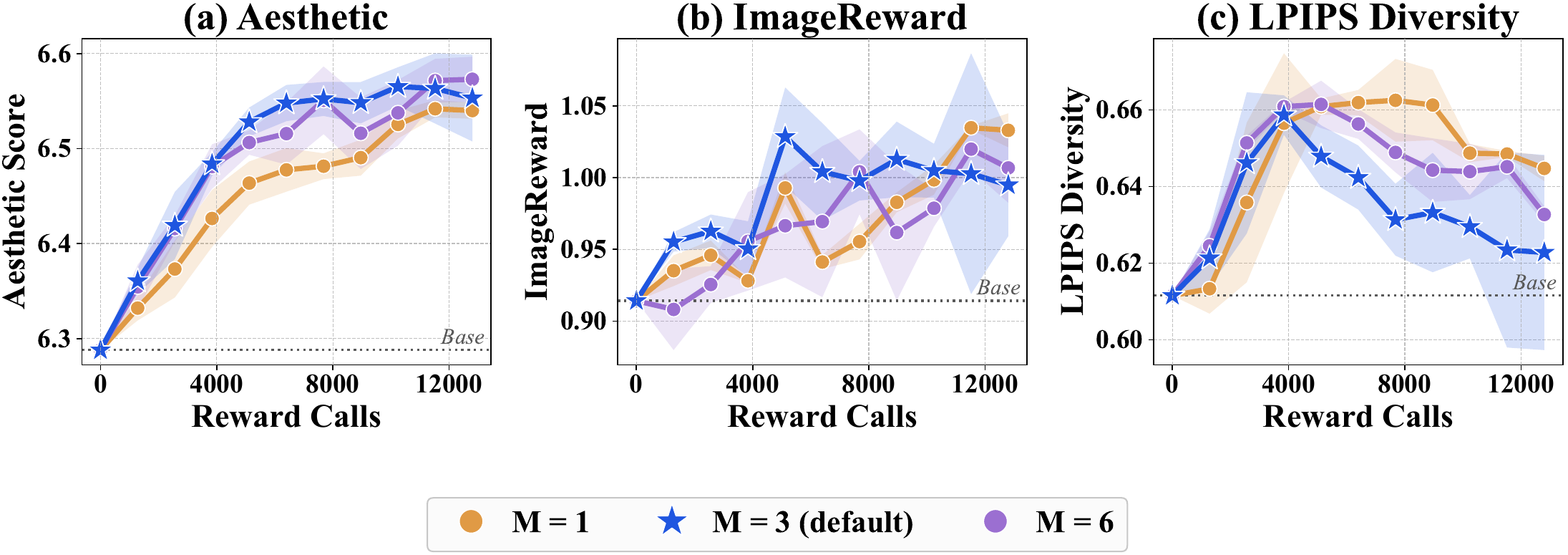}
\caption{\textbf{Ablation of the number of forward-noising pairs.}}
\label{fig:abl-mpairs}
\end{figure}

\paragraph{Number of forward-noising pairs.} \Cref{fig:abl-mpairs} varies the number $M$ of forward-noising pairs per endpoint in \Cref{eq:b-mc}. Increasing $M$ beyond $3$ brings no gain, so we use $M=3$ to limit the cost of the likelihood estimate.

\begin{figure}[H]
\centering
\includegraphics[width=\linewidth]{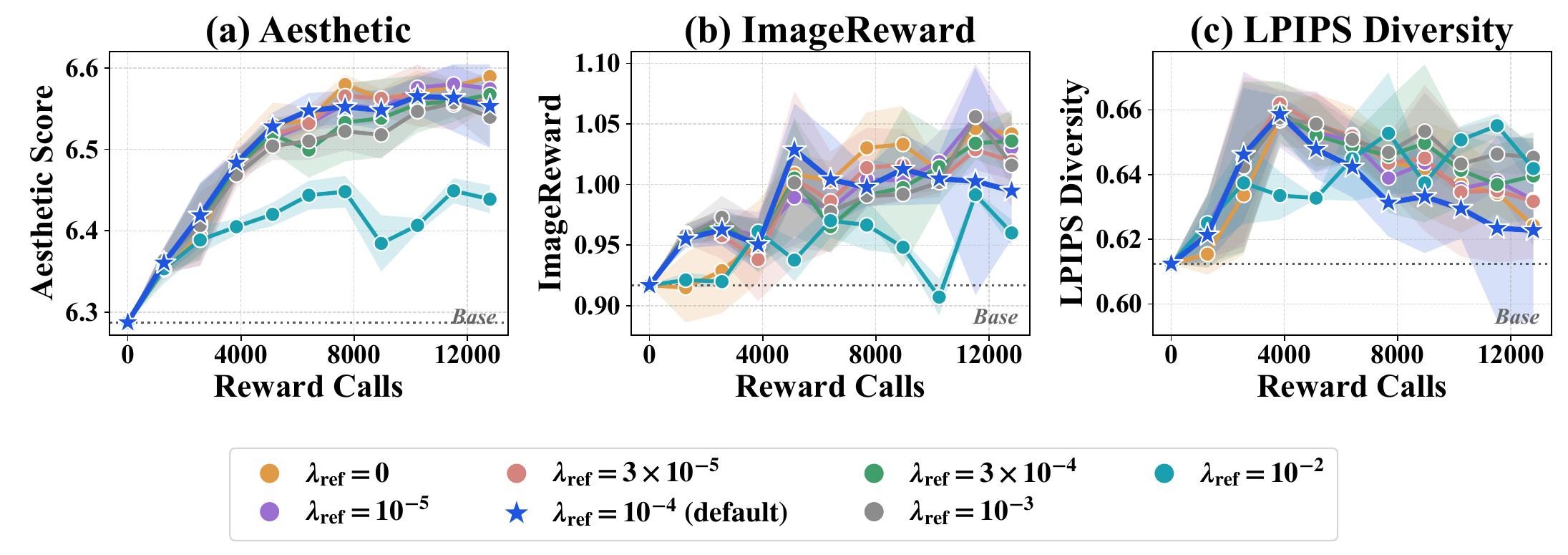}
\caption{\textbf{Effect of the regularization strength.}}
\label{fig:abl-reflambda}
\end{figure}

\begin{figure}[H]
\centering
\includegraphics[width=\linewidth]{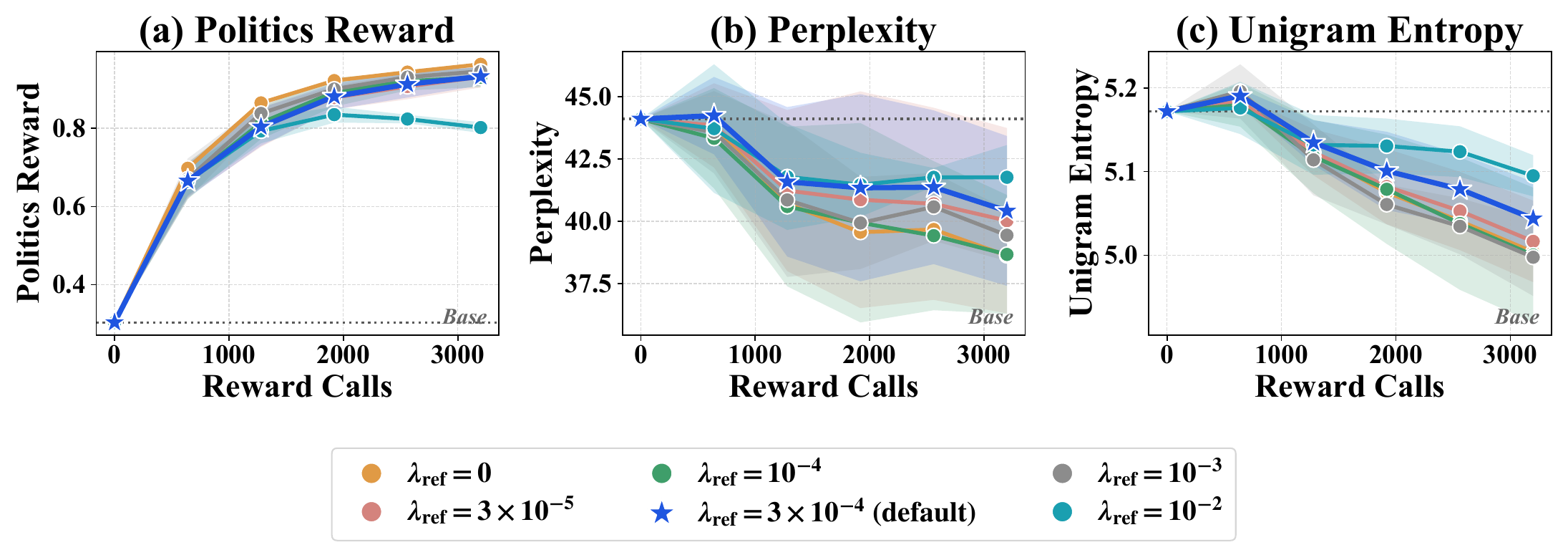}
\caption{\textbf{Effect of the regularization strength on language generation.} We additionally conduct language experiments to examine the effect of the regularization strength in another domain, using the politics topic of the language generation task with ELF (\Cref{sec:exp-language}).}
\label{fig:abl-reflambda-lang}
\end{figure}

\paragraph{Effect of the regularization strength.} \Cref{fig:abl-reflambda,fig:abl-reflambda-lang} vary $\lambda_{\mathrm{ref}}$ from $0$ to $10^{-2}$, extending the sweeps in \Cref{tab:image-hp-grid,tab:lang-hp-grid}. As the regularization strength increases, the target reward rises more slowly and reaches a lower final value, while the sample quality stays closer to that of the pretrained model.

\begin{figure}[H]
\centering
\includegraphics[width=\linewidth]{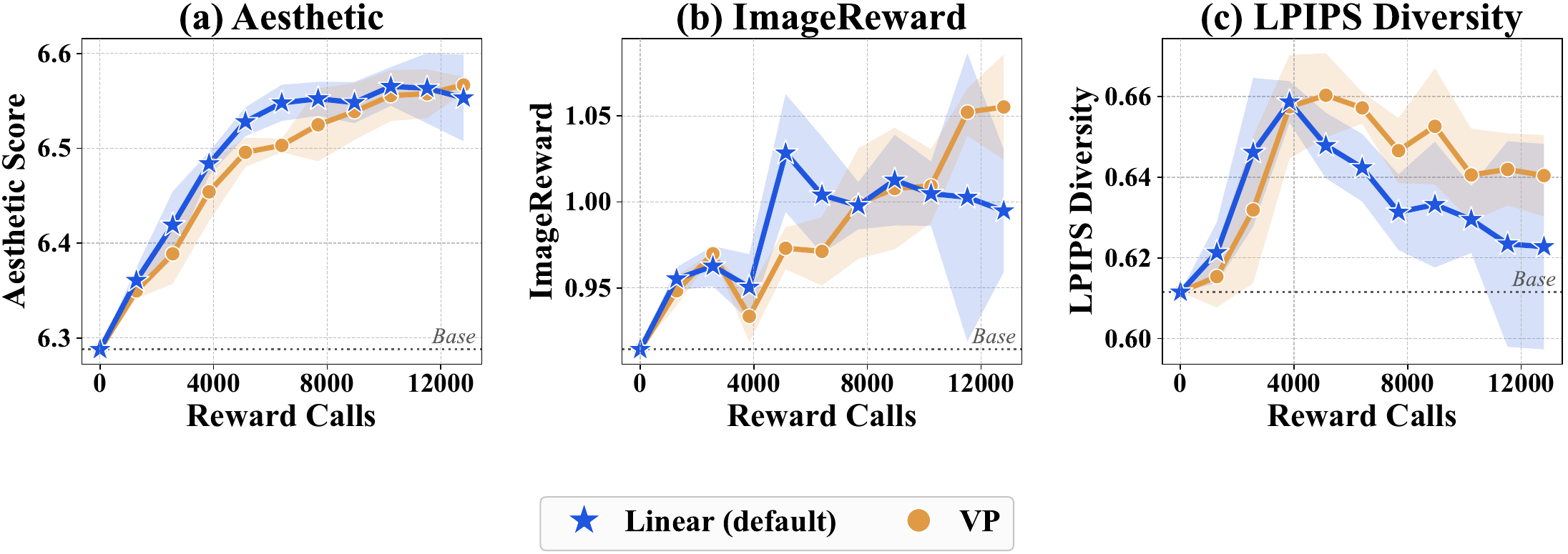}
\caption{\textbf{Ablation of the interpolation.}}
\label{fig:abl-interp}
\end{figure}

\paragraph{Interpolation.} \Cref{fig:abl-interp} replaces the linear interpolation of the reference flow with the variance-preserving interpolation of \Cref{app:reference-field}. The two interpolations perform similarly, so we use the linear interpolation.

\begin{figure}[H]
\centering
\includegraphics[width=\linewidth]{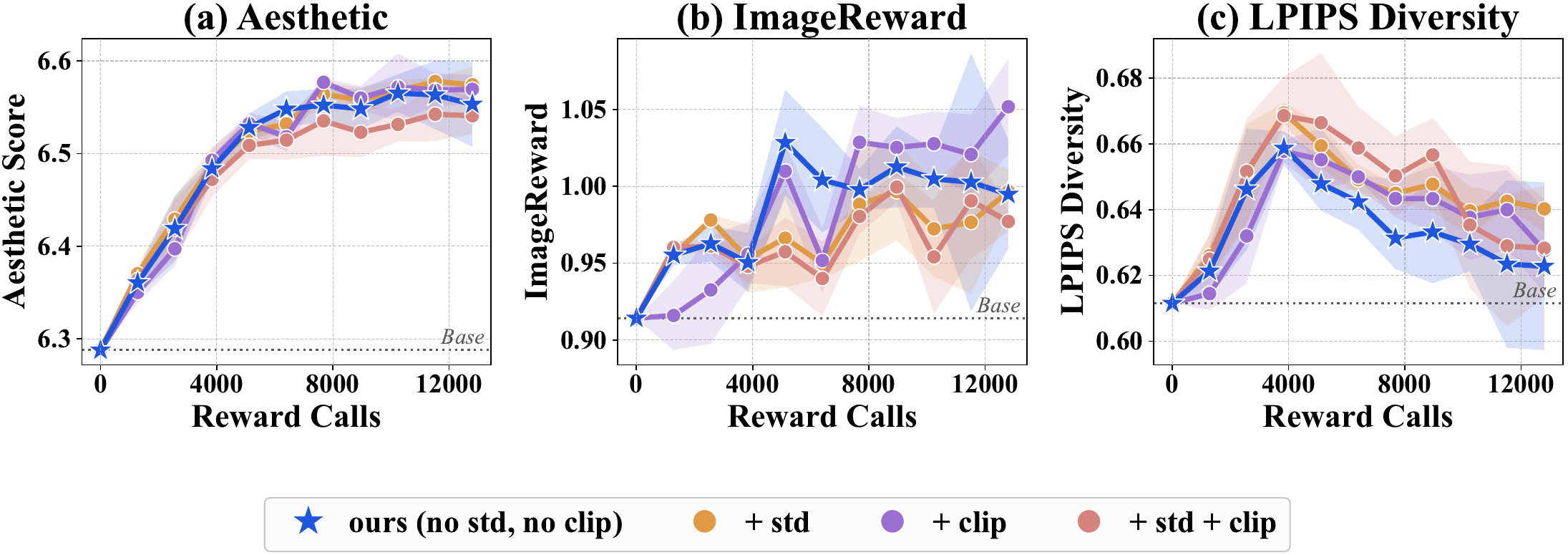}
\caption{\textbf{Ablation of the policy-gradient objective.}}
\label{fig:abl-pg}
\end{figure}

\paragraph{Policy-gradient objective.} \citet{choi2026rethinking} find that the choice of policy-gradient objective matters less than the likelihood estimator. For simplicity and fewer hyperparameters, NPO omits the within-group standard-deviation normalization of the advantage and the ratio clipping of GRPO. \Cref{fig:abl-pg} adds them back one at a time and together, which recovers the GRPO objective. None of them changes the results, so we keep the simpler update.

\subsection{Comparison across Regularization Strengths}
\label{app:reg-strength}

We report results with varying regularization strengths at the selected learning rate of each method. Darker markers indicate weaker regularization.

\begin{figure}[H]
\centering
\includegraphics[width=\linewidth]{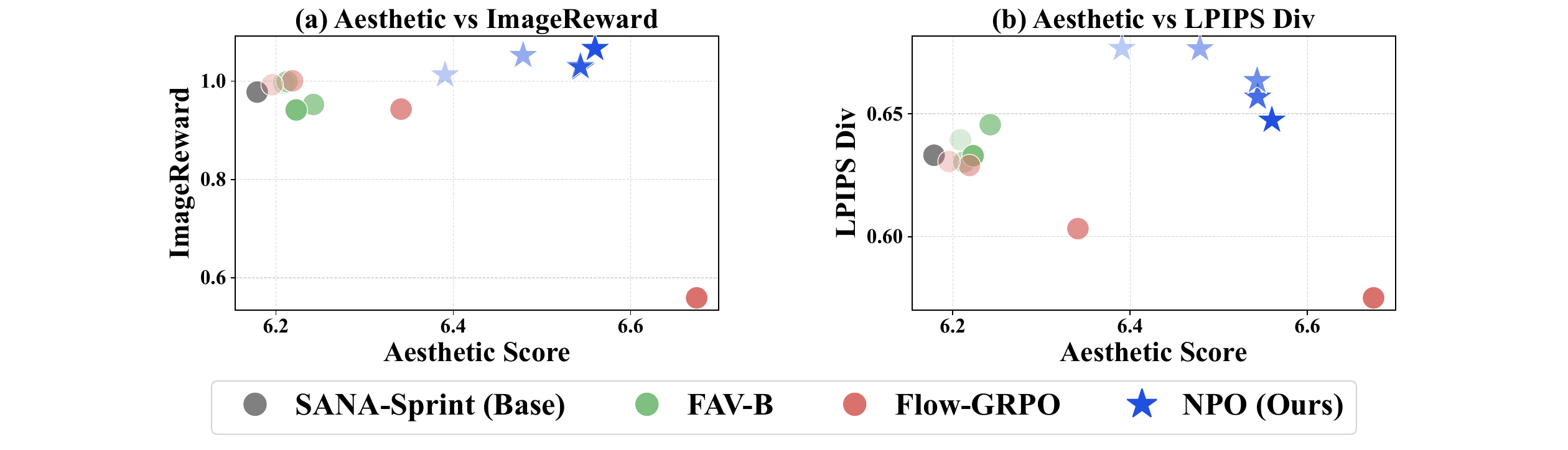}
\caption{\textbf{Reward-quality trade-off of NPO and direct fine-tuning baselines across regularization strengths on SANA-Sprint with four sampling steps.}}
\label{fig:reg-pareto-direct}
\end{figure}

\begin{figure}[H]
\centering
\includegraphics[width=\linewidth]{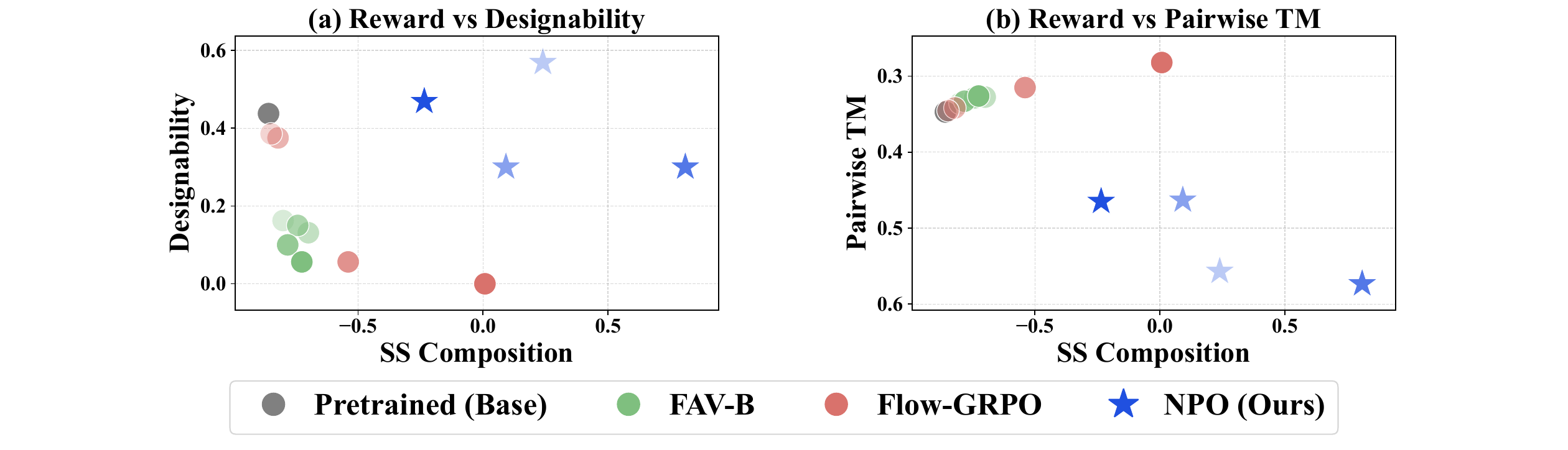}
\caption{\textbf{Reward-quality trade-off of NPO and direct fine-tuning baselines across regularization strengths on protein structure generation.}}
\label{fig:reg-pareto-bio}
\end{figure}

\paragraph{Direct fine-tuning.} We compare NPO with direct fine-tuning baselines across regularization strengths, evaluated at the same reward-call budget.
\begin{itemize}
\item \textbf{Image generation.} \Cref{fig:reg-pareto-direct} extends the aesthetic comparison of \Cref{fig:image_pareto_4step} to various regularization strengths. We vary $\beta \in \{10^{-3}, 10^{-2}, 10^{-1}, 1\}$ for Flow-GRPO and FAV-B, and $\lambda_{\mathrm{ref}} \in \{3\times10^{-5}, 10^{-4}, 3\times10^{-4}, 3\times10^{-3}, 10^{-2}\}$ for NPO. Across all regularization strengths, NPO achieves a better reward-quality trade-off than direct fine-tuning baselines.
\item \textbf{Protein structure generation.} \Cref{fig:reg-pareto-bio} extends panels (a) and (b) of \Cref{fig:bio_figure} to various regularization strengths. We vary $\beta \in \{4\times10^{-4}, 10^{-3}, 4\times10^{-3}, 10^{-2}\}$ for Flow-GRPO, $\beta \in \{10^{-4}, 3\times10^{-4}, 10^{-3}, 3\times10^{-3}, 10^{-2}\}$ for FAV-B, and $\lambda_{\mathrm{ref}} \in \{3\times10^{-5}, 10^{-4}, 3\times10^{-4}, 10^{-3}\}$ for NPO. Across all regularization strengths, direct fine-tuning baselines still exhibit substantial designability degradation as the target reward increases, while NPO maintains a better reward-designability trade-off.
\end{itemize}

\begin{figure}[H]
\centering
\includegraphics[width=\linewidth]{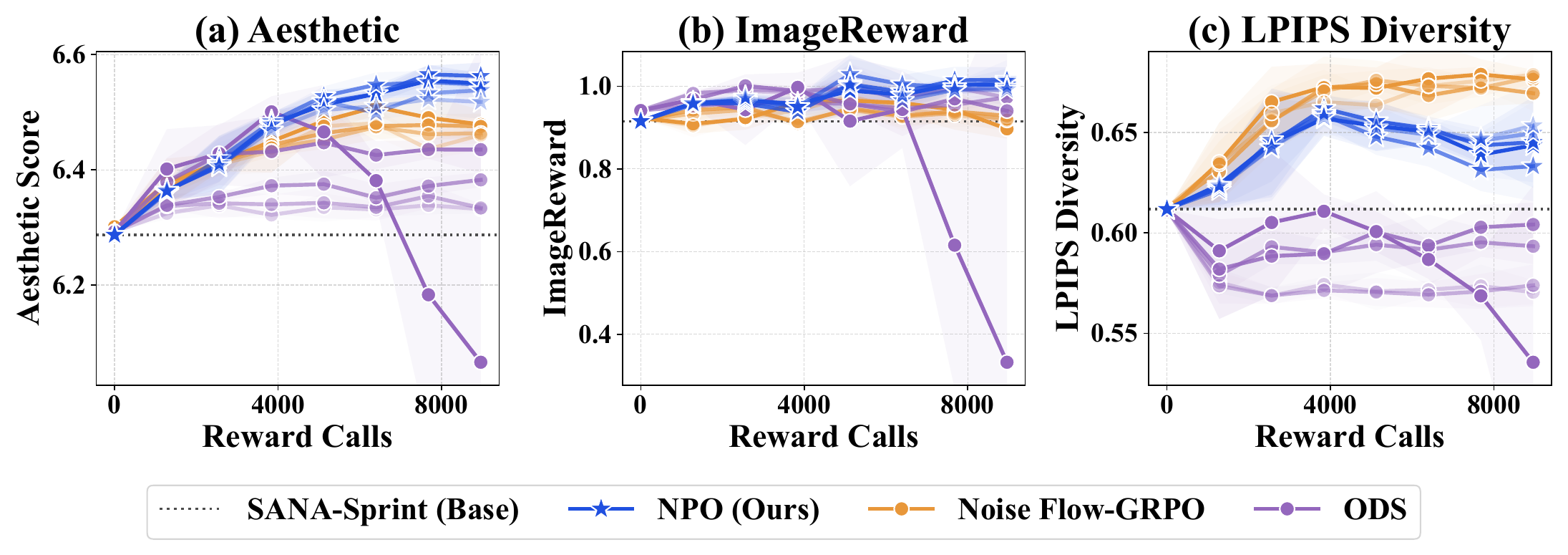}
\caption{\textbf{Reward learning curves of NPO and noise-policy learning baselines across regularization strengths on SANA-Sprint with one sampling step.}}
\label{fig:reg-curves-noise}
\end{figure}

\paragraph{Noise-policy learning.} \Cref{fig:reg-curves-noise} extends the one-step aesthetic comparison of \Cref{fig:image_noise_efficiency} to various regularization strengths, with $\lambda_{\mathrm{ref}} \in \{10^{-5}, 3\times10^{-5}, 10^{-4}, 3\times10^{-4}, 10^{-3}\}$ for NPO, $\beta \in \{4\times10^{-3}, 10^{-2}, 4\times10^{-2}, 10^{-1}\}$ for Noise Flow-GRPO, and $\beta \in \{50, 100, 500, 1000, 5000\}$ for ODS. Across all regularization strengths, NPO reaches comparable reward levels with fewer reward evaluations than trajectory-based noise-policy baselines.

\subsection{ELBO-Based Direct Fine-Tuning Baselines}
\label{app:elbo-direct}

By using ELBO-based likelihood estimation (\Cref{sec:method-elbo}) instead of trajectory-based likelihood estimation, NPO improves training efficiency in the noise space (\Cref{fig:image_noise_efficiency}). A natural question is whether ELBO-based likelihood estimation also benefits direct fine-tuning. AWM~\citep{xue2025advantage} and DiffusionNFT~\citep{zheng2026diffusionnft} follow this approach for multi-step flow models, approximating the likelihood with the flow-matching loss of the pretrained model.

We apply AWM and DiffusionNFT to the few-step generator under the additional assumption that its outputs can be treated as instantaneous velocities, and implement them as heuristic baselines.

\begin{figure}[H]
\centering
\includegraphics[width=\linewidth]{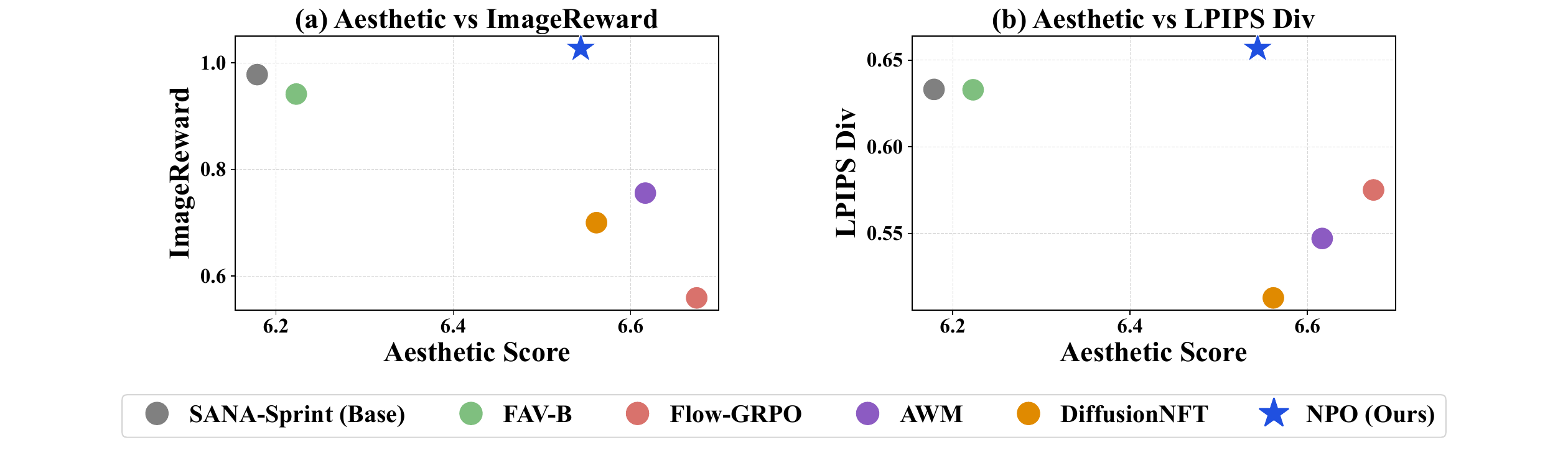}
\caption{\textbf{Reward-quality trade-off of NPO and direct fine-tuning baselines, including ELBO-based methods, on SANA-Sprint with four sampling steps.}}
\label{fig:elbo-direct}
\end{figure}

We fine-tune the attention layers of SANA-Sprint with LoRA, using the same rank, batch size, and reward-call budget as Flow-GRPO (\Cref{app:baseline-impl}). \Cref{fig:elbo-direct} extends the aesthetic comparison of \Cref{fig:image_pareto_4step} to include AWM and DiffusionNFT. Both show symptoms of reward over-optimization, especially diversity collapse. NPO maintains better sample quality with a comparable target reward, showing a better reward-quality trade-off than the ELBO-based direct fine-tuning baselines.

\clearpage
\section{Qualitative Comparison}
\label{app:qualitative}

\subsection{Image domain}
\label{app:qualitative-Image}

\begin{figure}[h]
\centering
\includegraphics[width=0.76\linewidth]{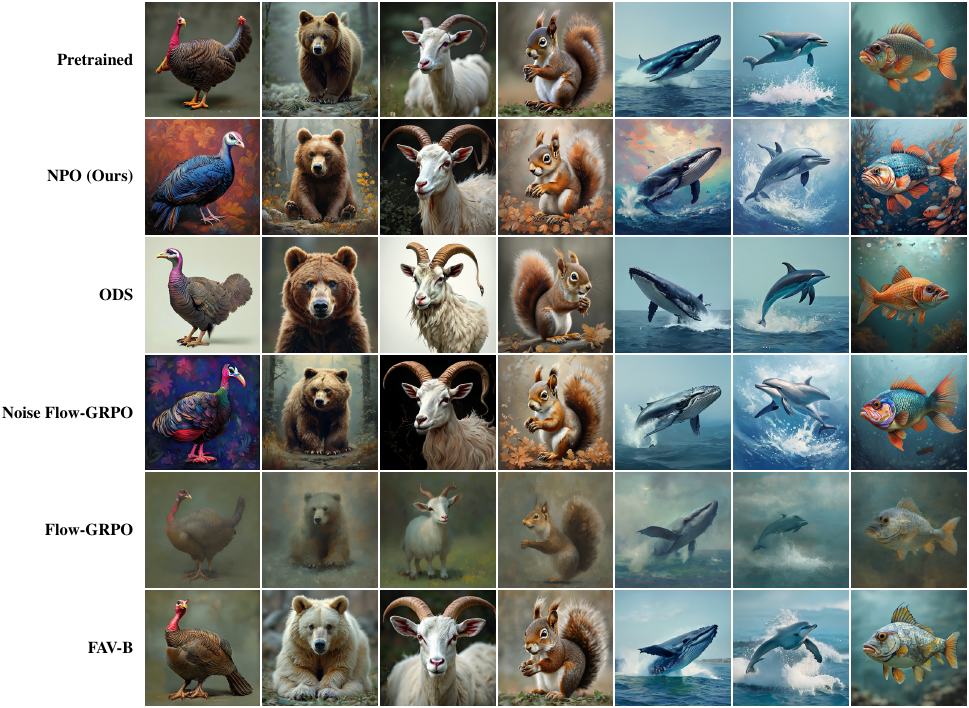}
\caption{\textbf{Qualitative comparison under the aesthetic reward.}}
\label{fig:qualitative-aesthetic}
\end{figure}

\begin{figure}[h]
\centering
\includegraphics[width=0.76\linewidth]{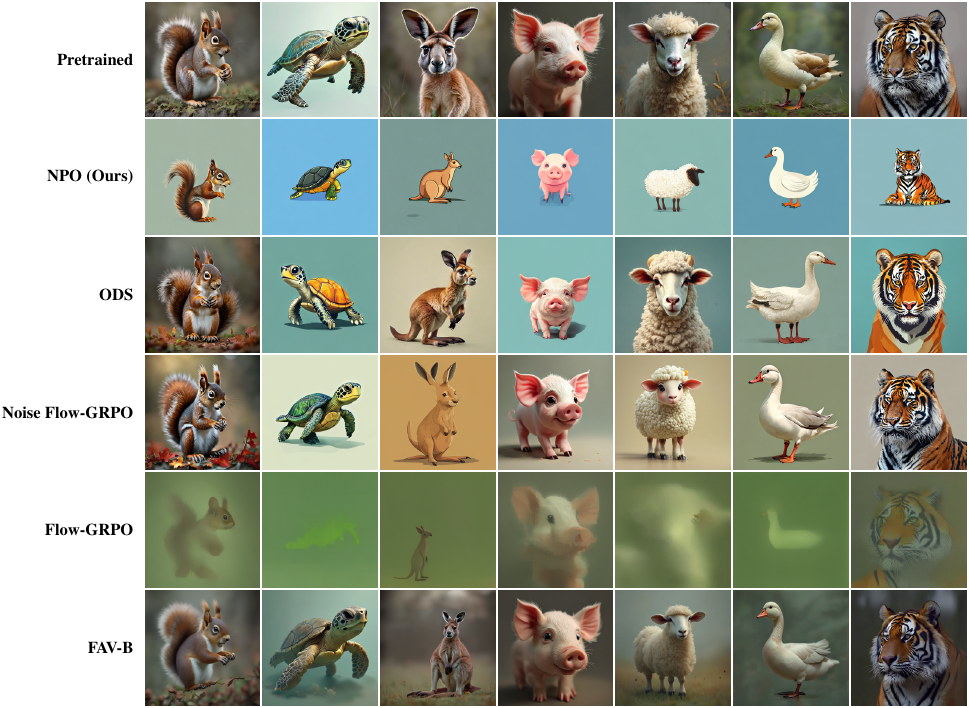}
\caption{\textbf{Qualitative comparison under the compressibility reward.}}
\label{fig:qualitative-compressibility}
\end{figure}

We show samples on SANA-Sprint with four sampling steps under the aesthetic reward (\Cref{fig:qualitative-aesthetic}) and the compressibility reward (\Cref{fig:qualitative-compressibility}). All samples are taken at a matched reward-call budget.

\clearpage
\subsection{Language domain}
\label{app:qualitative-lang}

We conduct a qualitative analysis on the topic-steering task and compare NPO with the direct fine-tuning methods on ELF. \Cref{fig:language_pareto} shows that the perplexity and unigram entropy of the direct fine-tuning methods decrease together as optimization proceeds. \Cref{fig:qualitative-lang-science,fig:qualitative-lang-politics} show how this collapse appears in the samples. We fix two initial noises and show their samples at three checkpoints of increasing training step. As optimization proceeds, the samples of the direct fine-tuning methods collapse into repetitions of a few tokens, which explains the simultaneous decrease in perplexity and unigram entropy.

\begin{figure}[h]
\centering
\scriptsize
\setlength{\tabcolsep}{5pt}
\renewcommand{\arraystretch}{1.12}
\begin{tabular}{@{}p{23mm}@{\hspace{1mm}}c@{\hspace{2mm}}p{112mm}@{}}
\toprule
\textbf{Base} & {\scriptsize 1} & As everyone has seen in the NFL, Tyry Johnson didn't appear, but he's one of the best runers in the NFL. We'll get to see from this. 22 DeJ McCrney,WR, WR, WR, Hardt : He's been a guy. \\
 & {\scriptsize 2} & If you're one of the most adventured fans on the game, we'll look at you a little to let you to see what Global Wring is up for. World Wring is one as one of the most engagingd games on the console that plays absolutely well about world order. \\
\midrule
\textbf{NPO} & {\scriptsize 1} & As everything happens in a pond, modern scientists can't exactly understand how energy is pumped by any thing that is moving through the air. But they might be able to do something from a piece of the lake into an atmosphere or through a piece inside ... \\
 & {\scriptsize 2} & This week, the nine month since Elson launched, we've talked with ToE and developers and producer Peter Chaleon to see what ORon is looking for. It's one of the big features that we've seen into the Nintendo- game slate. \\
\textbf{Flow-GRPO} & {\scriptsize 1} & In an interview released late Sunday, House President Donald Trump didn't say if he would place a tax on same-sex couples. His comments would be made public on July 5 at a seven-day music festival in Wilmington, Va. \\
 & {\scriptsize 2} & "We're not in a final situation, and we'ven't known what could happen next," Kailke said. While House Republican Republicans are voting against the seven-point vote, the Senate's Financeding Committee and the House Republican National Committee didn' ... \\
\textbf{FAV-B} & {\scriptsize 1} & Other people who live in the world who have't known or not never known knew other who have often two good friends who liked both met each other other others who who talked to other on Twitter and after each other often many times and talked with othe ... \\
 & {\scriptsize 2} & ``We're fortunate to be with our friends previously two previously and both met out to family and friends and others Derek Yel Walter said, Chief Executive and CEO of both day each two of our two two of our friends other close and known friends and pr ... \\
\midrule
\textbf{NPO} & {\scriptsize 1} & As everything happens in a physical world, academics can't exactly understand how energy is poured by an item that is moving through the air. But they might be able to push energy from a chunk of the mountain into an atmosphere or about a chunk below ... \\
 & {\scriptsize 2} & This week, the study was published in September in the Journal of Science, in the American Science Journal of Medicine. John Kailberg and several other Space Department researchers are working on the seven-wheel sensor from the NASA's Department Depa ... \\
\textbf{Flow-GRPO} & {\scriptsize 1} & In his remarks earlier on Thursday, US President Donald Trump didn't public, what he regarded as major threat to climate change in the Mediterranean. His remarks will be made public in 2017, including a seven-day wildlife festival in Edinburgh, Franc ... \\
 & {\scriptsize 2} & Last month, the day before President Obama announced his presidential election. He also visited the European House, China and on Thursday. He also outlined stronger climate change goals --- in his seven-day speech as the world's highest-elected preside ... \\
\textbf{FAV-B} & {\scriptsize 1} & \textcolor{collapse}{\itshape [1022 characters, all whitespace except ``\%'']} \\
 & {\scriptsize 2} & \textcolor{collapse}{\itshape [1021 characters, all whitespace except ``\%'']} \\
\midrule
\textbf{NPO} & {\scriptsize 1} & As everything happens nowadays in the physical world, investigators can't completely understand how chemical objects shaped a thing that is moving through the air. \\
 & {\scriptsize 2} & This study has been published in the journal Journal of Science Science. Research is supported by National Space Research Foundation and. John Cailski and other AEA researchers are working on the three-part study from the NASA's State Department of S ... \\
\textbf{Flow-GRPO} & {\scriptsize 1} & \textcolor{collapse}{\itshape [1024 characters in 1 unbroken token(s): ``.'' repeated]} \\
 & {\scriptsize 2} & \textcolor{collapse}{\itshape [1024 characters in 1 unbroken token(s): ``.'' repeated]} \\
\textbf{FAV-B} & {\scriptsize 1} & \textcolor{collapse}{\itshape [1051 characters, all whitespace except ``gone arrested arrested arreste'']} \\
 & {\scriptsize 2} & \textcolor{collapse}{\itshape [1031 characters, all whitespace except ``gone gone'']} \\
\bottomrule
\end{tabular}

\caption{Generated text samples on the \textbf{science} topic. \textcolor{collapse}{\itshape Red characters} indicate a collapsed sample with no content left to quote.}
\label{fig:qualitative-lang-science}
\end{figure}

\begin{figure*}[t]
\centering
\scriptsize
\setlength{\tabcolsep}{5pt}
\renewcommand{\arraystretch}{1.12}
\begin{tabular}{@{}p{23mm}@{\hspace{1mm}}c@{\hspace{2mm}}p{112mm}@{}}
\toprule
\textbf{Base} & {\scriptsize 1} & As everyone has seen in the NFL, Tyry Johnson didn't appear, but he's one of the best runers in the NFL. We'll get to see from this. 22 DeJ McCrney,WR, WR, WR, Hardt : He's been a guy. \\
 & {\scriptsize 2} & If you're one of the most adventured fans on the game, we'll look at you a little to let you to see what Global Wring is up for. World Wring is one as one of the most engagingd games on the console that plays absolutely well about world order. \\
\midrule
\textbf{NPO} & {\scriptsize 1} & As it appeared earlier this month, House President Donald Trump didn't express the reason he acted the right way about gun change in the country. The issue will be on display on TV, at a six-day music event in Detroit, Texas where he lost the 2016 el ... \\
 & {\scriptsize 2} & This week, the year-Julsith anniversary of Microsoft gathered with U\&C President and CEO Mike Melbeck to discuss what OpRon is working on its three-year project, the company's chief of development officer and vice president of computer design. \\
\textbf{Flow-GRPO} & {\scriptsize 1} & STTE: GETE: I don't don't believe it. He's a little bit odd. he's tricky. I'm attempting to get him get a bit cranky. He doesn't dance. he doesn't. He's very anxious. He think he's more delicate and he's so intelligent at something. \\
 & {\scriptsize 2} & Tese: I'm perseted [I'm really so intrigued]. What does that perspective Peter how would you get ask? Tese: I get. I'm one of the two. He's a very sophisticated player. I he's competent. he's a very basic rate. He's very remarkably sophisticated. \\
\textbf{FAV-B} & {\scriptsize 1} & "All the happens in the batting is very well kept and working extremely and he's a great guy that is efficient at the speed and I'll be able to play and at great level in the defensive and is able to quickly as he handles the punch and that's the how ... \\
 & {\scriptsize 2} & "We' are thrilled to be able to play and we're really looking forward to see him and see how he will and see what NBA he is looking at. He's one of the best stars from all on the world and player in the game really very well and overall he has a pers ... \\
\midrule
\textbf{NPO} & {\scriptsize 1} & As in the past three months, House Speaker Pe Carter didn't publicly confirm that President Trump did a good job decreasing gun change in the country. \\
 & {\scriptsize 2} & On Thursday, the EU-Omitsieur Presidenty Peen visited the White House with Secretary of State John Peilson to discuss what Global New Energy is doing after its three-day speech between the EU's chief of the staff and the secretary of National Securit ... \\
\textbf{Flow-GRPO} & {\scriptsize 1} & Rocker: In a way Stefan Franzens can't explain it. It's just this little thing that is flowing in the air. We're being able to pog into a trench that spills into an ocean of oil that ges into the ocean. \\
 & {\scriptsize 2} & Tese: We're crusing out what we're running the archaeological to looking the fossil and Peter Grelstein to see what Gresen is working at. These's one of the ones we're developing a piece of software that runs spatial graph maps. \\
\textbf{FAV-B} & {\scriptsize 1} & \textcolor{collapse}{\itshape [1023 characters, all whitespace except ``a'']} \\
 & {\scriptsize 2} & \textcolor{collapse}{\itshape [1023 characters, all whitespace]} \\
\midrule
\textbf{NPO} & {\scriptsize 1} & As it appeared earlier this month, House Speaker Rand Paul didn't say the way he stepped up how to reduce economic growth in the country. But things would be noticeable on screen, after a five-year restaurant production slump and 30 years after he le ... \\
 & {\scriptsize 2} & This Wednesday, about seven hours after Overselection President Donald Trump, visited the White House with Secretary of State John Caillke to discuss what Seranon is talking about. \\
\textbf{Flow-GRPO} & {\scriptsize 1} & Wimbledon Pour Pour Pour Pour Pour Pour impression impression impression Wimbledon impression impression impression impression impression impression impression impression impression impression impression impression impression impression impression im ... \\
 & {\scriptsize 2} & Pour Pour Pour Pour Pour Pour Pour Pour Pour impression impression impression impression impression impression impression impression impression impression impression impression impression impression impression impression impression impression impress ... \\
\textbf{FAV-B} & {\scriptsize 1} & \textcolor{collapse}{\itshape [337 tokens, 1 distinct: ``a'' repeated]} \\
 & {\scriptsize 2} & \textcolor{collapse}{\itshape [360 tokens, 1 distinct: ``a'' repeated]} \\
\bottomrule
\end{tabular}

\caption{Generated text samples on the \textbf{politics} topic. \textcolor{collapse}{\itshape Red characters} indicate a collapsed sample with no content left to quote.}
\label{fig:qualitative-lang-politics}
\end{figure*}

\subsection{Protein domain}
\label{app:qualitative-bio}

\begin{figure}[h]
\centering
\includegraphics[width=\linewidth]{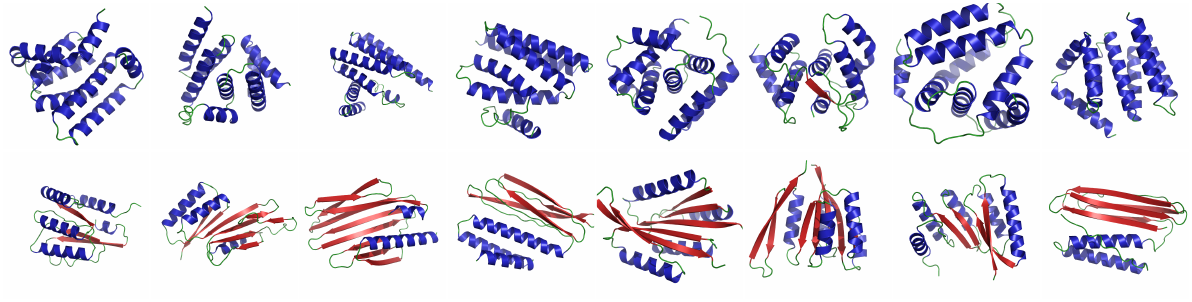}
\caption{\textbf{Qualitative comparison on protein structure generation.} Top: backbones generated by the pretrained generator. Bottom: backbones generated by NPO from the same initial noise. $\alpha$-helices are drawn in \textcolor[HTML]{2727C8}{blue}, $\beta$-sheets in \textcolor[HTML]{C82727}{red}, and coils in \textcolor[HTML]{27A027}{green}.}
\label{fig:bio-qualitative}
\end{figure}

We show protein backbones generated by NPO on the protein structure generation task. NPO optimizes the SS composition reward, which favors diverse secondary-structure compositions with additional emphasis on $\beta$-sheet content (\Cref{sec:exp-bio-setup}). \Cref{fig:bio-qualitative} compares the backbones that the pretrained generator and NPO generate from the same initial noise. The backbones of the pretrained generator consist mostly of $\alpha$-helices, whereas the backbones of NPO contain substantially more $\beta$-sheet content while keeping well-formed structures.

\end{document}